\documentclass[final,5p,times,twocolumn]{elsarticle}

\usepackage{amsmath,amssymb,amsfonts}
\usepackage{graphicx}
\usepackage{adjustbox}
\usepackage{tabularx}
\graphicspath{ {./images/} }
\usepackage{color,soul}
\usepackage{subcaption,booktabs}
\usepackage{multirow}
\usepackage{caption}
\usepackage{bm}
\usepackage[framemethod=tikz]{mdframed}
\usepackage{ multirow} % for borders and merged ranges
\usepackage{changepage,threeparttable} % for wide tables
\usepackage{ragged2e}
\usepackage{array}

\usepackage{natbib}
\usepackage[pagebackref=true,breaklinks=true,letterpaper=true,colorlinks,bookmarks=false, citecolor=Blue]{hyperref}
\usepackage{framed}
\usepackage{soul}
\usepackage{microtype}
\usepackage{pagenote}
\journal{Computers in Biology and Medicine}

\begin{document}

\begin{frontmatter}



\title{Leveraging Different Learning Styles for Improved Knowledge Distillation in Biomedical Imaging}


\author[label1]{Usma Niyaz}
\ead{usma.20csz0015@iitrpr.ac.in}

\author[label1]{Abhishek Singh Sambyal}
\ead{abhishek.19csz0001@iitrpr.ac.in}

\author[label1]{Deepti R. Bathula}
\ead{bathula@iitrpr.ac.in}

\affiliation[label1]{organization={Department of Computer Science and Engineering, Indian Institute of Technology Ropar},
            city={Rupnagar},
            postcode={140001},
            state={Punjab},
            country={India}}

\begin{abstract}

Learning style refers to a type of training mechanism adopted by an individual to gain new knowledge. As suggested by the VARK model, humans have different learning preferences, like Visual (V), Auditory (A), Read/Write (R), and Kinesthetic (K), for acquiring and effectively processing information. Our work endeavors to leverage this concept of knowledge diversification to improve the performance of model compression techniques like Knowledge Distillation (KD) and Mutual Learning (ML). Consequently, we use a single-teacher and two-student network in a unified framework that not only allows for the transfer of knowledge from teacher to students (KD) but also encourages collaborative learning between students (ML). Unlike the conventional approach, where the teacher shares the same knowledge in the form of predictions or feature representations with the student network, our proposed approach employs a more diversified strategy by training one student with predictions and the other with feature maps from the teacher. We further extend this knowledge diversification by facilitating the exchange of predictions and feature maps between the two student networks, enriching their learning experiences. We have conducted comprehensive experiments with three benchmark datasets for both classification and segmentation tasks using two different network architecture combinations. These experimental results demonstrate that knowledge diversification in a combined KD and ML framework outperforms conventional KD or ML techniques (with similar network configuration) that only use predictions with an average improvement of $2\%$. Furthermore, consistent improvement in performance across different tasks, with various network architectures, and over state-of-the-art techniques establishes the robustness and generalizability of the proposed model.

\end{abstract}


\end{frontmatter}

% \linenumbers
\pagenote{Accepted in Computers in Biology and Medicine}
\section{Introduction}
\label{sec:introduction}
{Undoubtedly, deep learning techniques perform exceptionally well in many domains, including medical diagnosis. However, the quest to achieve state-of-the-art performance has led to the development to highly complex neural networks. This restricts their application in many domains as they are computationally expensive to train and difficult to deploy. Specifically, most medical images are characterized by high resolution. While lightweight networks fail to capture the pixel-level spatial contextual information due to limited capacity,   highly parameterized models provide significantly superior performance. However, these giant models are impractical for real-world applications requiring resource-constrained edge or embedded devices deployment. Consequently, model compression is an active area of research that aims to reduce the configuration complexity of state-of-the-art deep networks to enable their deployment in resource-limited domains without significant reduction in performance \cite{deepcompress1,deepcompress2}.}

It is a misconception that only large and highly complex models achieve the best performance \cite{jimmy,howard2017mobilenets}. Many state-of-the-art approaches have shown that strategically designed lightweight models can provide similar performance \cite{buci}. It is now known that a significant percentage of nodes in these models are redundant, and pruning these connections minimally affects the performance \cite{NISP}. Several model compression techniques have been developed to simplify highly complex networks and substantially reduce the requirement for resources \cite{ali,fahad}. These include  Network Pruning \cite{np}, Quantization \cite{deepcompress2}, Low-Rank Matrix Approximation \cite{zhou}, Knowledge Distillation (KD) \cite{hinton}, Deep Mutual Learning (DML) \cite{dml},   and their variants. {Alternatively, scaling methods such as random search optimization \cite{random_search}, and EfficientNet \cite{tan2019efficientnet} aims to efficiently balance model complexity and performance by systematically adjusting various model dimensions.} The ultimate goal of all these techniques is to optimize the network configuration without compromising the model’s performance.

Network Pruning involves the elimination of nodes, weights, or layers that do not significantly influence performance. It speeds up the inference process by considering the parameters that matter the most for a specific task. Quantization-based approaches reduce the precision of numbers used to represent the neural network weights, and Low-rank approximation uses tensor decomposition to estimate the informative parameters of a network \cite{cheng}. Tiny machine learning is another fast-growing field that strives to bring the transformative power of machine learning to resource-constrained devices. It has shown significant potential for low-power applications in the vision and audio space \cite{tinyml}. 

Knowledge Distillation emerged as a potential and widely used model compression technique. It leverages the knowledge gained by a highly parameterized teacher network by sharing it with a lightweight student network such that the compact model approximates the performance of the teacher network. As an extension of the KD concept, {several alternative approaches have been proposed that leverage additional supervision from the pre-trained teacher model, particularly emphasizing the intermediate layers \cite{romeo,semckd,srrl}. Some of these approaches involve using spatial attention maps \cite{at}, while others explore the use of pairwise similarity patterns \cite{sp} or strive to maximize the mutual information between teacher and student features \cite{vid,SimKD}}. Other variants of KD, like Evolutionary Knowledge Distillation \cite{ekd} and Self Distillation \cite{sd}, were also introduced to enhance the performance of the student networks further. Knowledge distillation is well exploited in many healthcare applications for classification \cite{jou_m1,jou_m2, jou_b, jou_t,jou1,jou2}, localization \cite{loc1, loc2}, and segmentation \cite{seg,seg2}. It focuses on the problems of multimodality and small datasets \cite{modal1, kde-gan} and is often collaboratively used with other approaches to boost the performance of student networks\cite{sall,jou_p, jou_ssl}.

\begin{figure}[htbp]
\centering
\includegraphics[width=\linewidth]{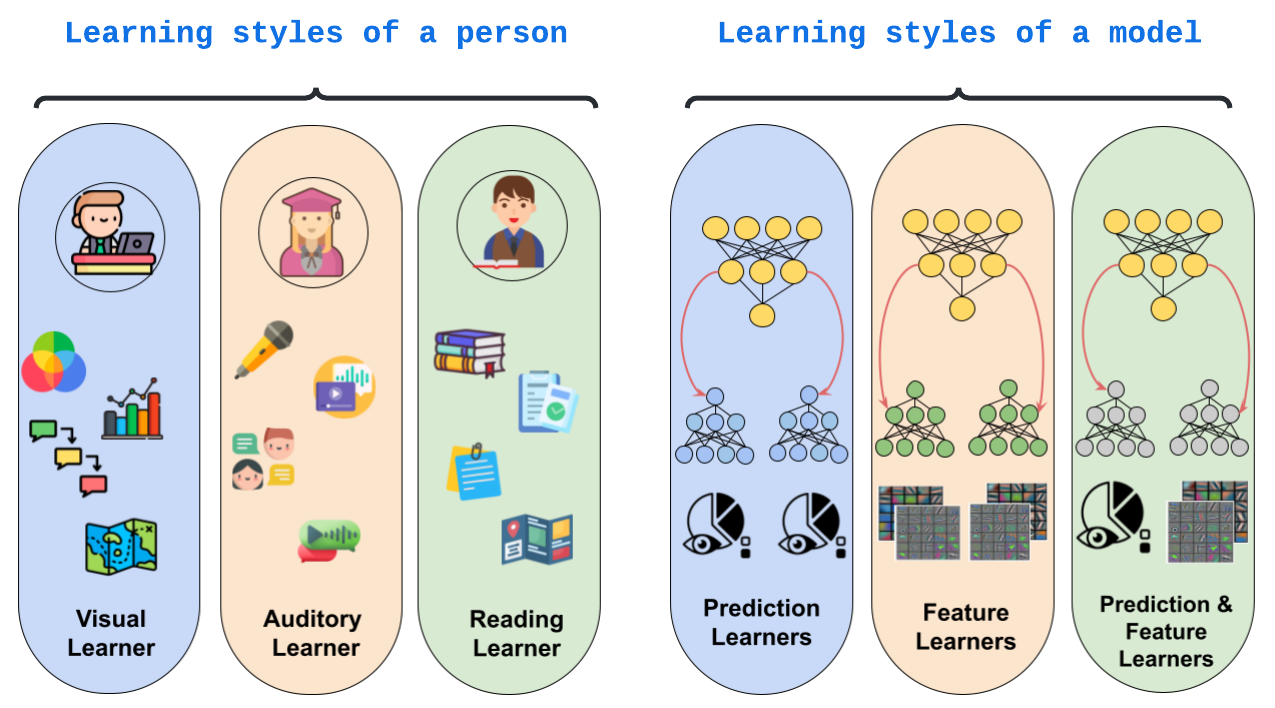}
\caption{ Illustration of a person's different learning style adopted for training the deep learning model. }
\label{fig:1}
\end{figure}

In these conventional techniques, knowledge transfer is typically limited to sharing predictions from a larger teacher network to a smaller student network (KD) or between two similar networks (ML). While alignment of predictions helps improve the performance, the logits have limited capacity to encapsulate complex insights. To address this limitation, Chen et al. \cite{kdfm}  introduced the use of feature maps in knowledge distillation. Due to their ability to capture high-level semantic information, feature maps were found to be more effective than logits for image classification.

In this work, we extend the knowledge distillation paradigm with the concept of diverse learning styles from classroom dynamics. A learning style refers to a type of training mechanism that an individual prefers to use to gain new knowledge. For example, as depicted in Figure \ref{fig:1}, the VARK model assumes four types of learners – Visual, Auditory, Reading \& Writing, and kinesthetic. Taking inspiration from this idea, we propose an enriched knowledge transfer protocol that incorporates the idea of different learning styles in terms of knowledge diversification. Consequently, we combine KD with ML in a single-teacher, multi-student framework to enable collaborative learning where the teacher imparts knowledge to the students, and the students also learn from each other.  Unlike conventional KD techniques, where the teacher shares the same knowledge with all the students, we propose to train individual student networks with varying forms of information from the teacher. Similarly, students exchange different types of information in the form of final predictions and intermediate layer features. 

The main contributions of this work are summarized as follows:

\begin{enumerate}

\item We propose an enhanced knowledge transfer protocol that incorporates the idea of different learning styles in terms of knowledge diversification.

%providing benefits to existing distillation approaches.

% \item Specifically, in a single-teacher, multi-student network that simulates classroom dynamics, the teacher network imparts knowledge to the student networks in diverse formats - such as predictions to one student and feature maps to another.
\item In a single-teacher, multi-student network that simulates classroom dynamics, the teacher network imparts knowledge to the student networks in diverse formats, such as predictions to one student and feature maps to another. Further, we enrich each student's learning process by facilitating the exchange of diversified knowledge among them.

\item Extensive experiments on metastatic tissue classification, brain tumor segmentation, and dermatological segmentation and classification tasks demonstrate the superiority of the proposed method over conventional knowledge distillation approaches.

\item Finally, the improvement afforded by the knowledge diversification strategy is attributed to the increased similarity of learned representations between higher layers of the teacher and student networks as measured by the Centered Kernel Alignment (CKA) metric.

\end{enumerate}

\begin{figure*}[htbp]
\centering
\captionsetup{labelfont=bf}
 \includegraphics[width=0.9\linewidth]{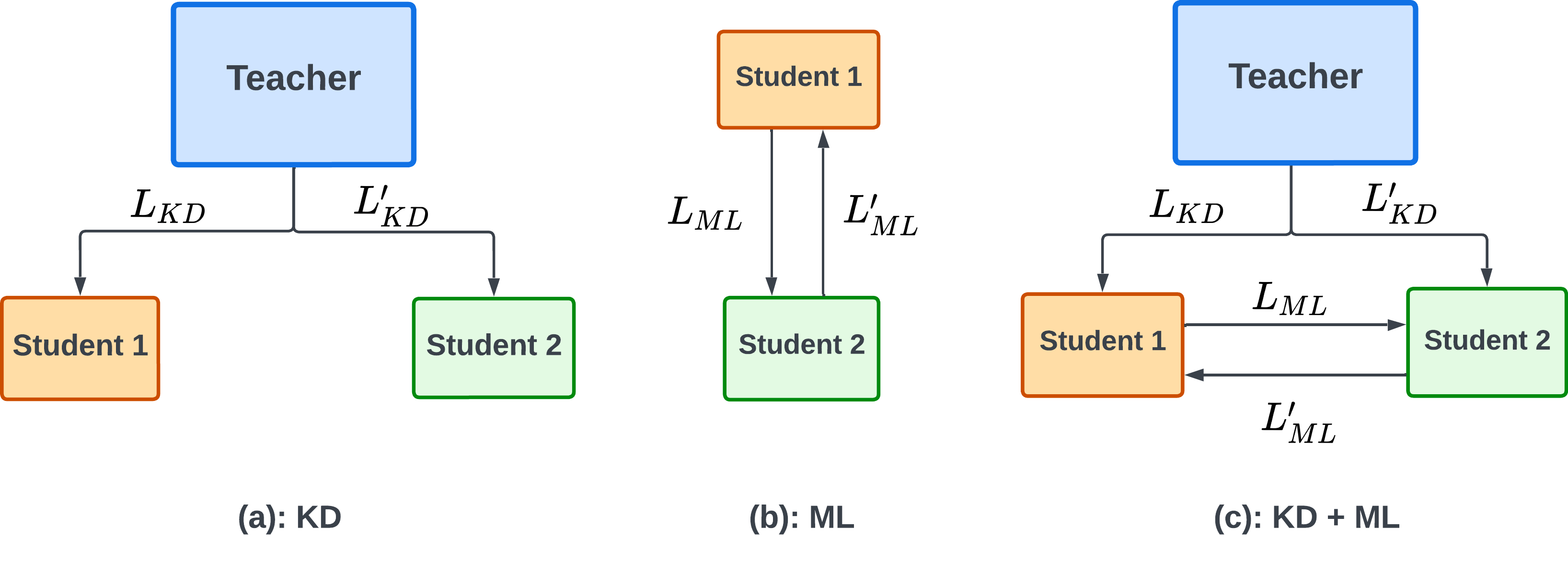}
\caption{\textbf{KD} - Transfers knowledge in terms of soft labels from a large, pre-trained teacher to a compact student network; \textbf{ML} - Both networks are considered as students for information exchange; \textbf{KD + ML} - Training students with knowledge from teacher network as well as from each other. {Conventionally, the same type of knowledge was shared with both student networks with $L_{KD} = L'_{KD}$ and $L_{ML} = L'_{ML}$. (The size of the block represents the complexity of a model, and color represents different architectures.)}}
\label{fig:2}
\end{figure*}

\begin{table*}[h]
% \resizebox{12cm}{!}{ 
% \captionsetup{labelfont=bf}
    \caption{Combinations of distillation techniques and learning strategies employed: (a) KD – Knowledge Distillation only, (b) ML – Mutual Learning only, (c) combined KD and ML; V1 – sharing of final predictions only, V2 – sharing of features only, and a diverse knowledge paradigm V3 – sharing of predictions and features together.}
    \begin{subtable}{.5\linewidth}
      \centering
        \caption{}
        \begin{tabular}{|l|c|c|c|}
        \hline \textbf{KD} & V1 &V2 &V3\\
            \hline
			T $\rightarrow$ $S_{1}$ & Predictions &  Features &  Predictions  \\
			T $\rightarrow$ $S_{2}$ & Predictions & Features &  Features  \\
			\hline
        \end{tabular}
    \end{subtable}%
    \begin{subtable}{.5\linewidth}
      \centering
        \caption{}
       \begin{tabular}{|l|c|c|c|}
			\hline \textbf{ML} & V1 &V2 &V3\\
			\hline
			$S_{1}$ $\rightarrow$ $S_{2}$ & Predictions &  Features &  Predictions  \\
		   $S_{2}$ $\rightarrow$ $S_{1}$ & Predictions & Features &  Feature  \\
			\hline
		\end{tabular}
      
    \end{subtable} 
    \begin{subtable}{1\linewidth}
      \centering
        \caption{}
        % \resizebox{11.8cm}{!}{
      		\begin{tabular}{|l|c|c|c|}
			\hline \textbf{KD + ML} & V1 &V2 &V3\\
			\hline
			T $\rightarrow$ $S_{1}$, $S_{2}$ $\rightarrow$ $S_{1}$ & Predictions, Predictions &    Features, Features & Features, Predictions  \\
			T $\rightarrow$ $S_{2}$, $S_{1}$ $\rightarrow$ $S_{2}$ & Predictions, Predictions &  Features, Features  & Predictions, Features  \\
			\hline
		\end{tabular}
      % }
    \end{subtable} 
    % }
    \label{tab:1}
\end{table*}

\section{Related work}

\subsection{Knowledge Distillation}
Knowledge Distillation \cite{hinton} is an approach introduced to transfer the knowledge in terms of probability outputs, $p_{i}$, from a  {complex, highly parameterized pre-trained teacher network $f(X,\phi)$ to a simple and compact student network $g(X,\theta)$ to achieve model compression while retaining the high performance of the teacher.} \\
Given a training set with $N$ samples 
${X}=\left\{\boldsymbol{x}_{i}\right\}_{i=1}^{N}$ with corresponding labels ${Y}=\left\{y_{i} \right\}_{i=1}^{N}$, the teacher network {$f(X,\phi)$}, is trained on the ground truth labels. The probabilistic output of a teacher network for a sample $x_{i}$  is defined as $p_{i}$ given by the extended softmax as:
\begin{equation}
p_{i}^{c} = \frac{e^{\mathbf{z}^{c}/T}}{\sum_{c=1}^C e^{\mathbf{z}^{c}/T}} \ \ \ for\ c=1,2,\dots,C
% p_{i}=\frac{\exp \left(z_{c'}/T\right)}{\sum_{c=1}^{C} \exp \left(z_{c}/T\right)}
\end{equation}
\noindent where $\mathbf{z}^{c}$ corresponds to the logits, $C$ is the number of classes, and $T$ is the temperature parameter to get a smoother output probability distribution of the classes. Generally, the objective function for the teacher network is the standard \emph{Cross-Entropy (CE) error} defined as:
\begin{equation}
L_{\phi}= L_{CE}\left({p}_{i} , {y}_{i}\right)=-{\sum_{i=1}^N(y_{i}\log(p_{i}) + (1 - y_{i})\log(1 - p_{i}))}
\end{equation}
Now, the student networks are trained on the combined loss of \emph{Cross-Entropy (CE)}, and \emph{Knowledge Distillation (KD)}, where the \emph{CE} helps the student networks to adhere to the ground truth labels and \emph{KD} assists them to align their learning with that of the teacher. Here, \emph{Kullback Leibler (KL)} divergence \cite{kl} is used for $L_{KD_p}$ to measure the correspondence between the teacher and student predictions $p_{i}$ and $s_{i}$ respectively as:
% are the knowledge distillation is attained by obtaining the weighted distillation loss over the probability output,  and $s_{i}$, of complex teacher and  distilled student model calculated by the extended softmax as:

\begin{equation}
\label{eqn:KL}
L_{KD_p} = D_{KL}\left(\boldsymbol{s}_{i} \| \boldsymbol{p}_{i}\right)=\sum_{i=1}^{N} {s}_{i}\left({x}_{i}\right) \log \frac{{s}_{i}\left({x}_{i}\right)}{{p}_{i}\left({x}_{i}\right)}
\end{equation}

\noindent Finally,  the loss function for the student network is the weighted $(\alpha) $ summation of the cross entropy $(L_{CE})$ and knowledge distillation $(L_{KD_p})$ terms: 
%Here, \emph{Kullback Leibler (KL)} divergence \cite{kl} is used for $L_{KD_p}$ to measure the correspondence between the teacher and student predictions $p_{i}$ and $s_{i}$ respectively.
\begin{equation}
\label{eqn:KD}
L_{\theta}= \alpha \ L_{CE}\left({s}_{i} , {y}_{i}\right) + (1-\alpha) \ L_{KD_p}\left({s}_{i} , {p}_{i}\right)
\end{equation}
{where hyperparameter $\alpha$ is used to balance the contributions of the hard target loss ($L_{CE}$) and soft target loss ($L_{KD_{p}}$) during the distillation process for each student}. The knowledge can be transferred in an online or offline manner from the teacher to the student networks. In offline knowledge distillation (KD (off)), training is done in two steps; first, the teacher network is pre-trained on a dataset, and then the knowledge is distilled to train the student network, whereas in online knowledge distillation (KD (on)) \cite{kdcl}, both teacher and student networks are trained simultaneously in a single training step.

\begin{figure*}[htbp]
% \captionsetup{labelfont=bf}
\centering
 \includegraphics[width=\linewidth]{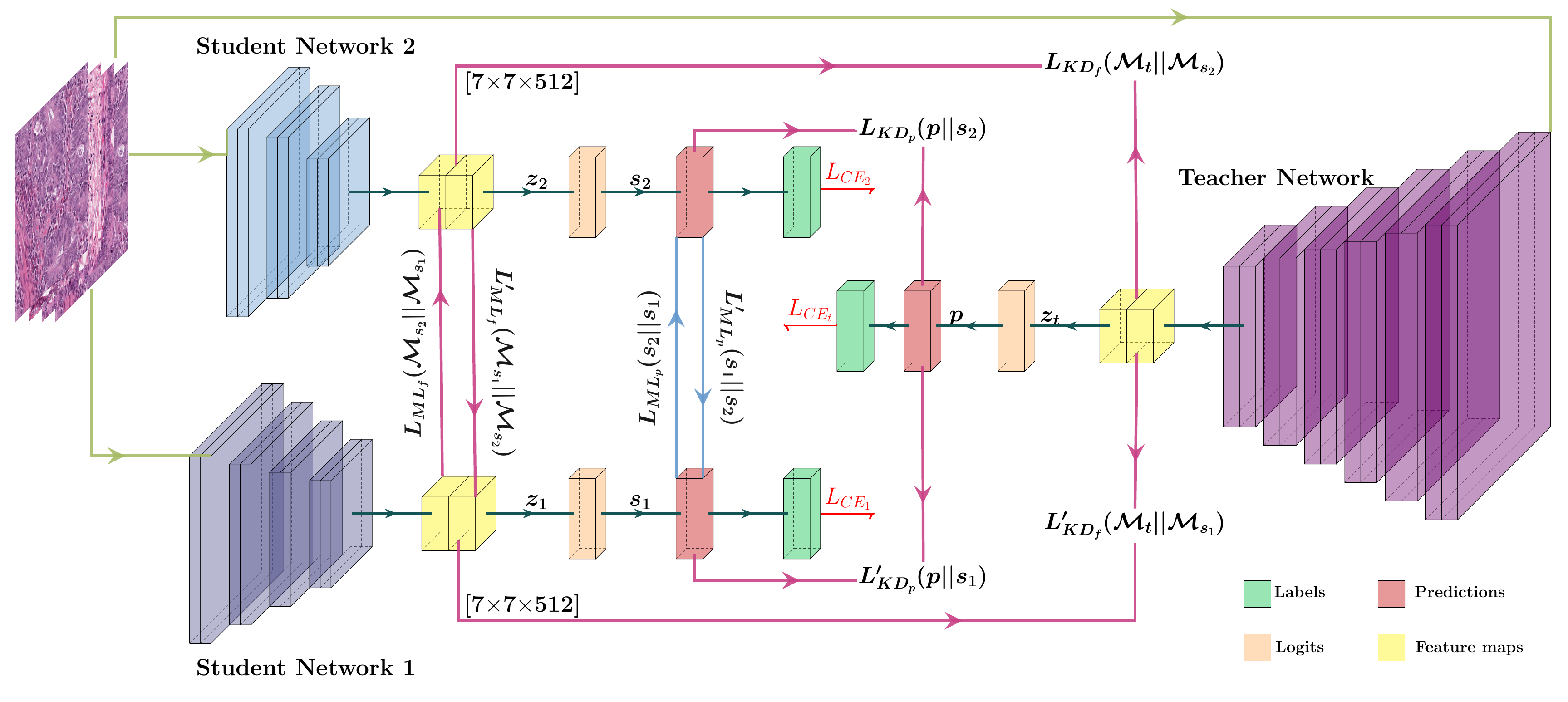}
\caption{Overview of our proposed model that combines Knowledge Distillation (KD) with Mutual Learning (ML) in a single-teacher, two-student framework, leveraging a diversified knowledge-sharing strategy in the classification task. The loss terms $L$ and $L'$ represent the respective losses for each student, while $L_{KD_{p}}$ and $L_{KD_{f}}$ capture the knowledge distillation losses between the teacher and students over predictions and features, respectively. Similarly, $L_{ML_{p}}$ and $L_{ML_{f}}$ denote the mutual learning losses between the two students over predictions and features, respectively. Here ResNet50 is used as a teacher network and ResNet18 as a student network  }
\label{fig:10}
\end{figure*}

\subsection{Deep Mutual Learning}
Unlike knowledge distillation, mutual learning \cite{dml} is a two-way sharing of information as both networks are treated as student networks. {Both the students can be of the same or different configuration.
}They teach each other collaboratively throughout the entire training process. The loss functions $L_{\theta_{1}}$ and $L_{\theta_{2}}$ for the two student networks {$g_{1}(X,\theta_{1})$} and  {$g_{2}(X,\theta_{2})$} respectively are defined as

\begin{equation}
\label{eqn:ML}
\begin{array}{l}
L_{\theta_1}= L_{CE}\left(\boldsymbol{s}_1 ,\boldsymbol{y}\right)+ L_{ML_p}\left(\boldsymbol{s}_1 , \boldsymbol{s}_2\right) \\
\\
L_{\theta_2}=L_{CE}\left(\boldsymbol{s}_2 ,\boldsymbol{y}\right)+ L_{ML_p}\left(\boldsymbol{s}_2 , \boldsymbol{s}_1\right)
\end{array}
\end{equation}
% where $s_{1}$ and $s_{2}$ are the predictions of the student network 1 and 2 respectively,
where $\boldsymbol{s}_{k}$, $k \in \{1,2\}$ is the predictions of the \emph{kth} student network, and similar to \emph{KD}, $L_{ML_p}$ defined as the  \emph{Mutual Learning loss} is the \emph{KL} divergence between the predictions of two students. Therefore, $L_{ML_p}(\boldsymbol{s}_{k'},\boldsymbol{s}_{k}) = D_{KL}(\boldsymbol{s}_{k'} || \boldsymbol{s}_{k})$, where the \emph{KL} distance from $\boldsymbol{s}_{k'}$ to $\boldsymbol{s}_{k}$ is computed using equation \ref{eqn:KL}.

%\begin{equation}
%D_{KL}\left(\boldsymbol{s}_{k'} \| \boldsymbol{s}_{k}\right)=\sum_{i=1}^{N} {s}_{k'}\left({x}_{i}\right) \log \frac{{s}_{k'}\left({x}_{i}\right)}{{s}_{k}\left({x}_{i}\right)}
%\end{equation}

\section{Method}
\subsection{Knowledge Distillation and Mutual Learning (KDML) with Knowledge Diversification }
Figures \ref{fig:2} (a) and (b) depict the standard distillation techniques used for knowledge distillation and mutual learning. Extension of KD to a single teacher and multiple student networks is also quite standard. Recently, \cite{UNiyaz} proposed a combination of KD and ML, where in addition to sharing information by the teacher, students also exchange information, as shown in Figure \ref{fig:2} (c). In such configurations, the ensemble of student networks is used for the final prediction. In the current work, we investigate the benefit of knowledge diversification in four different distillation techniques- (i) Offline KD-only framework as shown in Figure \ref{fig:2} (a) (ii) Online KD-only framework (iii) ML-only framework with two students as shown in Figure \ref{fig:2} (b), and (iv) combined KD + ML frameworks with one teacher and two students as shown in Figure \ref{fig:2} (c).

Leveraging the idea that different learning styles can improve the understanding of learners, we propose to train individual student networks with different information from the teacher. The teacher shares final predictions with one student and intermediate layer features with another. Similarly, in the ML framework, students engage in information exchange, sharing predictions and intermediate-layer features with one another. To underscore the significance of knowledge diversification, we train each of the above-mentioned distillation techniques with three different information-sharing strategies (V1, V2, and V3), as shown in Table \ref{tab:1}. Furthermore, we use student networks with identical architectures to emphasize the influence of different learning styles.

We hypothesize that diversifying knowledge has the potential to enhance the efficacy of the three established distillation techniques, namely KD (off), KD (on), and ML. We anticipate that the most optimal performance can be achieved by combining the KD and ML configuration with the ensemble of two student networks trained with distinct information from the teacher and promoting the exchange of diverse knowledge between them. 

\subsection{Application in Classification Task}

  In our approach, which uses a combined KD + ML configuration with a knowledge diversification strategy to distill knowledge from the teacher while facilitating mutual learning among students, we define the respective loss functions as follows:
\begin{equation}
\label{eqn:KDMLV3C1}
% \begin{array}{l}
\begin{split}
L_{\theta_1}=\alpha \ L_{CE}\left(\boldsymbol{s}_1 , \boldsymbol{y}\right) & + \beta \ L_{KD_f}\left(\mathcal{M}_{s_1} , \mathcal{M}_{t}\right)\\
& + \gamma \ L_{ML_p}\left(\boldsymbol{s}_1 , \boldsymbol{s}_2\right)
\end{split}
\end{equation}

\begin{equation}
\label{eqn:KDMLV3C2}
% \begin{array}{l}
\begin{split}
L_{\theta_2} =\alpha' \ L_{CE}\left(\boldsymbol{s}_2 , \boldsymbol{y}\right) & + \beta' \ L'_{KD_p}\left(\boldsymbol{s}_2,\boldsymbol{p} \right)\\ 
& + \gamma' \ L'_{ML_f}\left(\mathcal{M}_{s_2} , \mathcal{M}_{s_1}\right)
\end{split}
\end{equation}

\noindent {where $L'_{KD_p}$ and $L_{ML_p}$ represent the same loss terms based on predictions as defined in equations \ref{eqn:KD} and \ref{eqn:ML} respectively. To encourage knowledge diversification, we introduce two supplementary loss terms, $L_{KD_f}$ and $L'_{ML_f}$. These loss terms are constructed based on features shared by the teacher, represented as $\mathcal{M}_{t}$, and the other student, denoted as $\mathcal{M}_{s_k}$. We define the feature map-based loss function as the \emph{Mean Square Error (MSE)} between the feature maps of the corresponding networks. In general, for \emph{n} number of feature maps, the \emph{MSE} between two feature maps is defined as $\frac{1}{n} \sum_{i=1}^{n} (\hat{\mathcal{M}_i}-\mathcal{M}_i)^2$.} {The detailed depiction of our proposed approach with a knowledge diversification strategy is shown in Figure \ref{fig:10}, and the network architecture details are given in Table \ref{tab:7} in the supplementary material. Refer to Table \ref{tab:13} (supplementary material) for a comprehensive view of the different loss terms involved in the distillation methods with various learning styles}.

Using Equations \ref{eqn:KDMLV3C1} and \ref{eqn:KDMLV3C2}, we can derive the loss functions for KD-only and ML-only configurations with knowledge diversification by setting weighing parameters $\gamma = \gamma’ = 0$ and $\beta = \beta’ = 0$ respectively. As each student is learning from different information, we use separate weighing parameters for individual terms of the loss function and optimize them using grid search.  For a test sample $x_{i}$, we consider the ensemble classification probability, $\hat{s}(x_i)$, as the highest probability for a particular class across all student network predictions. This is given as ${\hat{s}(x_{i})}=max\{{{s}_k(x_{i})}\}_{k=1}^{K}$.

{Although sharing predictions is a common practice, a more detailed explanation is required for the sharing of feature information. With the assumption that the last layers of deep neural networks encode high-level semantic information, we propose to use the output of the teacher network's last convolution layer as feature information to share with student networks. However, to enable this knowledge transfer, it is necessary to ensure that the student network's convolutional block has an output feature map with dimensions matching the teacher network's. This ensures an effective transfer of knowledge between the two networks.
In KD configurations, where the teacher and student networks do not have layers with matching dimensions to share or compare the feature map information, an additional convolutional block can be added to the teacher network with an output dimension to match that of the student. This ensures the compactness of the student networks. A similar approach can also be adopted for ML configurations with non-identical student networks.}

  %\textcolor{red}{For a convolutional block with kernel size $M$ and a number of filters $D$, the dimension of the output feature map is given as $\mathcal{M} = M \times M \times D$.}
  %\begin{mdframed}[hidealllines=true,backgroundcolor=yellow!20]{However, to enable this knowledge transfer, it is necessary for the student network's convolutional block to have an output feature map size of M = M × M × D, matching that of the teacher network. This ensures an effective transfer of knowledge between the network.}\end{mdframed}

\subsection{Application in Segmentation Task}

In general, U-Net \cite{unet} is the preferred and most commonly used architecture for image segmentation. To capture essential feature information for sharing with student networks, we employ the output feature map from the initial convolution layer of the teacher's decoder network, as depicted in Figure \ref{fig:11} in the supplementary material. It was found empirically that this layer, being close to the encoder, contains valuable semantic information. It helps generate more precise predictions when combined with the information from previous layers of the encoder through skip connections. Unlike classification task that uses cross-entropy loss ($L_{CE}$), we use a combination of \emph{Focal loss (FL)} \cite{fl} and \emph{Dice loss (DL)} \cite{dl}, defined as $L_{FD}$,  to train the networks with ground truth segmentation labels. The loss function, $L_{FD}$ between the predicted $\boldsymbol{\hat{g}}$ and ground truth mask $\boldsymbol{g}$, is defined as:
\begin{equation}
\begin{split}
L_{FD}(\boldsymbol{\hat{g}},\boldsymbol{g}) & = \underbrace{-\sum_{i=1}^N\left(1-\hat{g}_i\right)^\tau \log \left(\hat{g}_i\right)}_{\text {Focal Loss }}\\
& + \underbrace{1-\frac{2 \sum_{i=1}^N \hat{g}_i g_i}{\sum_{i=1}^N \hat{g}_i^2+\sum_{i=1}^N g_i^2}}_{\text {Dice Loss }}
\end{split}
\end{equation}
where ${\hat{g}_{i}}$ and ${g}_{i}$ are the corresponding predicted probability and ground truth, and $\tau$ is the focusing parameter that enables the model to prioritize 
 hard examples during training. The $L_{FD}$ is most effective for segmentation as \emph{ Dice loss} handles unequal distribution of foreground-background elements, whereas \emph{Focal loss} alleviates the problem of class imbalance by using a down-weighting mechanism to reduce the influence of easy examples that are well-classified and focus on hard examples that require additional attention during training. 
 % This approach can improve the overall performance of the model, particularly in cases where one class is significantly underrepresented.
Consequently, the loss functions for the segmentation task using KD + ML configuration with a knowledge diversification are formulated as follows:
\begin{equation}
\label{eqn:KDMLV3S1}
\begin{split}
L_{\theta_1}=\alpha \ L_{FD}\left(\boldsymbol{s}_1 , \boldsymbol{y}\right)+ \beta \ L_{KD_f}\left(\mathcal{M}_{s_1} , \mathcal{M}_{t}\right) \\ +
\gamma \ L_{ML_p}\left(\boldsymbol{s}_1 , \boldsymbol{s}_2\right)
\end{split}
\end{equation}

\begin{equation}
\label{eqn:KDMLV3S2}
% \begin{array}{l}
\begin{split}
L_{\theta_2}=\alpha' \ L_{FD}\left(\boldsymbol{s}_2 , \boldsymbol{y}\right) + \beta' \ L'_{KD_p}\left(\boldsymbol{s}_2,\boldsymbol{p} \right) \\ + 
\gamma' \ L'_{ML_f}\left(\mathcal{M}_{s_2} , \mathcal{M}_{s_1}\right)
\end{split}
\end{equation}
Similar to the classification task, using Equations \ref{eqn:KDMLV3S1} and \ref{eqn:KDMLV3S2}, we can derive the segmentation loss functions for knowledge diversified KD-only and ML-only configurations by setting $\gamma = \gamma’ = 0$ and $\beta = \beta’ = 0$ respectively. Finally, the ensemble prediction mask is calculated as the union of individual student predictions, ${\hat{y}(x_{i})}=\bigcup_{i=1}^n\{{{s}_k(x_{i})}\}_{k=1}^{K}$.

\section{Experiments}
\subsection{Dataset details}
For the classification task, we used the Histopathologic Cancer Detection Dataset \cite{class} consisting of 220k images for identifying metastatic tissue in histopathologic scans of lymph node sections. Due to limited computational resources, a balanced subset of 20,000 histological images was randomly selected from the repository for our experiment. {Each image represents a patch of size $96 \times 96$ extracted from histopathologic scans of lymph node sections, annotated with a binary label indicating presence of metastatic tissue.}
For the segmentation task, we chose the  Low-Grade Gliomas (LGG) Segmentation Dataset \cite{segment}. {This dataset consists of 3929 MR brain images obtained from The Cancer Imaging Archive (TCIA) \cite{tcia}. Each image has a resolution of $256 \times 256$ and a corresponding manual FLAIR abnormality segmentation mask. The datasets for both tasks are split into training, validation, and testing with a 75:10:15 ratio.}
{We have conducted additional experiments using the HAM10000 dataset for both (a) Lesion Classification and (b) Lesion segmentation tasks. A more detailed analysis of these experiments is provided in the supplementary material (Section A.6). }

\subsection{Implementation details}
%We used the ResNet50 and ResNet18 architectures for the classification task as our teacher and student models, respectively. For segmentation, state-of-the-art UNET architecture is utilized where the backbones for teacher and student networks are the same ResNet50 and ResNet18 models. For this particular configuration, the combined student networks reduce $15\%$ in the number of parameters compared to the teacher network.  
{
We used two different network architecture combinations for our single teacher and two student configurations – (a) ResNet50 as a teacher with ResNet18 as students and (b) ResNet50 as a teacher and MobileNet as students. While these networks can be used directly for the classification task, they were used as backbones for the teacher and student networks in the U-Net framework for the segmentation task. ResNet50, ResNet18, and MobileNet are all pre-trained models equipped with approximately 26 million, 11 million, and 4.8 million parameters, respectively. Consequently, the ensemble of two ResNet18 students and the ensemble of two MobileNet students provide model compression of $15\%$ and $63\%$, respectively, when compared to the ResNet50 teacher network.}

{The feature-sharing aspect of our proposed classification model requires that the dimensions of the feature maps of teacher and student networks match. The feature map of the last convolutional layer of ResNet50 is $7 \times 7 \times 2048$, ResNet18 and MobileNet have dimensions of $7 \times 7 \times 512$ and $7 \times 7 \times 1024$, respectively. To resolve this discrepancy, an additional convolutional block is introduced in the teacher network containing $1 \times 1$ convolutional layer with the number of filters to match the student – 512 for ResNet18 and 1024 for MobileNet. For the segmentation task, as the feature maps extracted from the first convolutional layer in the decoder of both the teacher and student networks have the same dimensions ($16 \times 16 \times 256$), no additional modifications are required.}

%\textcolor{blue}{ However, for the segmentation task, the feature maps of the first convolutional layer in the decoder for both teacher and student networks have the same spatial dimension of [16 $\times$ 16] and a depth of 256 channels. Therefore, there was no need to introduce any additional convolutional layers in the decoder }

{We employed standard data augmentation techniques \cite{data_aug} widely recognized in the field of machine learning. These include horizontal flips, vertical flips, rotation, transpose, shift, and scale, along with normalization to introduce variations in orientation, position, scale, and intensity. By increasing the diversity, these techniques help counteract over-fitting and increase the robustness of the models.} The optimal values of all the hyper-parameters for different distillation techniques and learning strategies were identified using a grid search. All the models were trained with Adam optimizer with a learning rate of $0.0001$, batch size as $8$, and the temperature parameter $T$ as 2. These parameters are selected empirically. We report our models' average and standard deviation of 3 runs for a more robust evaluation.

\subsection{Experimental setup}

We conducted a comprehensive set of experiments to evaluate the robustness and generalizability of our model. Firstly, we compared four different distillation techniques – (i) ML, (ii) KD (on), (iii) KD (off), and (iv) Combined KD and ML (KD + ML).  Secondly, we explored the performance of these four techniques using three different learning strategies – V1 (predictions only), V2 (features only), and a diverse knowledge paradigm, V3 (both predictions and features). Different combinations of distillation and learning strategies lead to a total of 12 different variants. {The performance of these 12 variants was evaluated for classification and segmentation tasks on three standard datasets. Finally, these experiments were repeated using two different teacher-student network architecture combinations, ResNet50-ResNet18 and ResNet50-MobileNet.}

\subsection{Evaluation metrics}
To assess the effectiveness of the proposed 
knowledge distillation approach, we used accuracy for the classification models and Intersection-over-Union (IoU) and F1-score for the segmentation task as defined below:

\begin{equation}
Accuracy = \frac{TP+TN}{TP+TN+FP+FN}
\end{equation}

\begin{equation}
F1 = \frac{2*TP}{2*TP+FP+FN}
\end{equation}

\begin{equation}
IoU = \frac{TP}{TP+FP+FN}
\end{equation}

Where TP, TN, FP, and FN represent the number of True Positive, True Negative, False Positive, and False Negative predictions, respectively. Accuracy measures the proportion of correct predictions to the total number of predictions, and IoU
assesses the overlap between a model’s predicted and the ground truth segmentations.
% \end{mdframed}

%############################## 
% Baseline table
%#####################################
\begin{table}[h]
% \captionsetup{labelfont=bf}
\centering
% \captionsetup{font=small}
\caption{Performance comparison of baseline models on metastatic tissue classification and LGG Segmentation.}

% \resizebox{8.2cm}{!}{
\begin{tabular}{|c||c||c|c|c|}

\hline
\multirow{2}*{Model}&{Classification} &\multicolumn{2}{c|}{{Segmentation}} \\
\cmidrule{2-4}
 &{Accuracy} &{IOU} &{F-score} \\
\hline
ResNet50 &$94.35\pm0.763$ &$76.86\pm0.791 $&$86.93\pm0.537$ \\
\hline
ResNet18 &$94.07\pm0.436$ & $75.06\pm0.792$& $85.13\pm 0.561$\\
\hline
MobileNet &$91.38\pm0.491$ & $68.15\pm0.287$& $80.24\pm 0.225$\\
\hline
\end{tabular}
% }
\label{tab:2}
\end{table}

%############################## 
% classification table
%##################################

\begin{table}[h!]
% \captionsetup{labelfont=bf}
\centering
% \captionsetup{font=small}
      \caption{Performance comparison of ResNet50-ResNet18 for classification accuracy for Histopathologic Cancer Detection dataset using four different distillation techniques: ML - Mutual Learning, KD (on) - online Knowledge Distillation, KD (off) - offline Knowledge Distillation, and KD + ML – combined KD \& ML; and three different learning strategies -  V1 (predictions only), V2 (features only) and a diverse knowledge paradigm, V3 (both predictions and features).}
    \begin{subtable}{\linewidth}
    \centering
  \resizebox{9.2cm}{!}{       
        \begin{tabular}{|c|c|c|c|}
        \hline
         %  \multirow{2}*{ML}&{V1} & V2 & V3   \\
         % & ($\alpha=0.2,\alpha'=0.2$) & ($\alpha=0.2,\beta=0.2$)   & ($\alpha=0.1,\beta=0.2$) \\      
         \multirow{5}*{ML}&{V1} & V2 & V3   \\
         & ($\alpha=0.2,\alpha'=0.2$)   & ($\alpha=0.2,\alpha'=0.2$) & (${\alpha=0.1},\alpha'=0.2$)\\
         & ($\beta=0,\beta'=0$)   & ($\beta=0,\beta'=0$)& (${\beta=0},\beta'=0$)  \\
         & ($\gamma=0.8,\gamma'=0.8$)   & ($\gamma=0.8,\gamma'=0.8$) & (${\gamma=0.9},\gamma'=0.8$)\\
         \hline
         \textbf{S1} & $94.25\pm0.042$ &$92.52\pm0.487$ &$95.04\pm0.095$ \\
         \hline
\textbf{S2} &$94.15\pm0.442$  &$93.52\pm0.219$ &$94.89\pm0.245$\\
\hline
\textbf{Ensemble} &$94.22\pm0.155$  &$94.19\pm0.106$ &$\mathbf{95.36\pm0.239}$\\
\hline
 \hline
         
    \end{tabular}
    }
    
    \label{tab:31}
    \end{subtable}
% \vspace{2.5mm}
    \begin{subtable}{\linewidth}
    \centering
\resizebox{9.2cm}{!}{ 
            \begin{tabular}{|c|c|c|c|}
        \hline
        \multirow{5}*{KD (off)}&{V1} & V2 & V3   \\
         
         & ($\alpha=0.2,\alpha'=0.2$)   & ($\alpha=0.2,\alpha'=0.2$) & (${\alpha=0.2},\alpha'=0.2$)\\
         & ($\beta=0.8,\beta'=0.8$)   & ($\beta=0.8,\beta'=0.8$)& (${\beta=0.8},\beta'=0.8$)  \\
         & ($\gamma=0,\gamma'=0$)   & ($\gamma=0,\gamma'=0$) & (${\gamma=0},\gamma'=0$)\\
         \hline
        \textbf{T} &$94.35\pm 0.763$  &$94.35\pm0.763$ &$94.35\pm0.763$\\
         \hline
\textbf{S1} &$94.02\pm0.576$ &$94.32\pm0.176$ &$94.59\pm0.127$  \\
 \hline
\textbf{S2} &$94.31\pm0.134$ &$94.25\pm0.245$  &$94.87\pm0.530$\\
 \hline
\textbf{Ensemble} &$94.75\pm0.954$ &$94.68\pm0.176$ &$\mathbf{95.45\pm0.490} $ \\
\hline
 \hline
         
    \end{tabular}
    }
    \label{tab:32}
    \end{subtable}{}

    % \vspace{2.5mm}
    \begin{subtable}{\linewidth}
    \centering
\resizebox{9.2cm}{!}{ 
            \begin{tabular}{|c|c|c|c|}
        \hline
 \multirow{5}*{KD (on)}&{V1} & V2 & V3   \\
          & ($\alpha=0.2,\alpha'=0.2$)   & ($\alpha=0.2,\alpha'=0.2$) & (${\alpha=0.1},\alpha'=0.2$)\\
         & ($\beta=0.8,\beta'=0.8$)   & ($\beta=0.8,\beta'=0.8$)& (${\beta=0.9},\beta'=0.8$)  \\
         & ($\gamma=0,\gamma'=0$)   & ($\gamma=0,\gamma'=0$) & (${\gamma=0},\gamma'=0$)\\
         \hline
\textbf{T} &$94.43\pm0.353$  &$94.15\pm1.312$ &$95.43\pm0.707$ \\
\hline
\textbf{S1} &$94.74\pm0.191$  &$93.23\pm0.869$ &$95.75\pm1.079$ \\
\hline
\textbf{S2} &$94.55\pm0.281$ &$94.38\pm0.912$ &$95.23\pm0.438$ \\
\hline
\textbf{Ensemble} &$95.18\pm0.162$  &$95.06\pm0.776$ &$\mathbf{96.15\pm0.707}$ \\
 \hline
 \hline
         
    \end{tabular}
   } 
       \label{tab:33}
    \end{subtable}{}

    % \vspace{2.5mm}
    \begin{subtable}{\linewidth}
    \centering
\resizebox{9.2cm}{!}{ 
            \begin{tabular}{|c|c|c|c|}
        \hline
         \multirow{5}*{KD + ML}&{V1} & V2 & V3   \\
         & ($\alpha=0.1,\alpha'=0.2$)   & ($\alpha=0.2,\alpha'=0.2$) & (${\alpha=0.2},\alpha'=0.4$)\\
         & ($\beta=0.45,\beta'=0.4$)   & ($\beta=0.4,\beta'=0.4$)& (${\beta=0.4},\beta'=0.3$)  \\
         & ($\gamma=0.45,\gamma'=0.4$)   & ($\gamma=0.4,\gamma'=0.4$) & (${\gamma=0.4},\gamma'=0.3$)\\
         
         \hline
\textbf{T} &$93.21\pm2.341$  &$94.46\pm0.309$ &$95.75\pm0.756$ \\
\hline
\textbf{S1} &$94.37\pm0.883$ &$95.46\pm0.487$ &$95.85\pm0.968$ \\
\hline
\textbf{S2} &$94.71\pm0.518$ &$95.06\pm0.353$ &$95.37\pm0.353$  \\
\hline
\textbf{Ensemble} &$95.25\pm0.360 $&$96.15\pm0.219 $ &$\textcolor{blue}{\bm{96.68\pm0.584}}$ \\
\hline

 \hline
         
    \end{tabular}
    }
       \label{tab:34}
    \end{subtable}{}

\label{tab:3}
\end{table}

%############################## 
% Segmentation table
%##################################
\begin{table*}[t]
% \captionsetup{labelfont=bf}
\renewcommand{\arraystretch}{1.2}
% \captionsetup{font=small}
\centering
        \caption{Performance comparison of U-Net with ResNet50-ResNet18 encoder for segmentation task using LGG dataset with IoU and F-score metrics using four different distillation techniques: ML - Mutual Learning, KD (on) - online Knowledge Distillation, KD (off) - offline Knowledge Distillation, and KD + ML – combined KD \& ML; and three different learning strategies - V1 (predictions only), V2 (features only) and a diverse knowledge paradigm, V3 (both predictions and features). }
    \begin{subtable}{\linewidth}
    \centering
    	% \hspace*{-100cm} 
\resizebox{13cm}{!}{       
        \begin{tabular}{|c|c|c|c|c|c|c|}
        \hline
\multirow{5}*{ML} &\multicolumn{2}{c|} {V1} &\multicolumn{2}{c|} {V2} &\multicolumn{2}{c|} {V3} \\
         & \multicolumn{2}{c|}{$\alpha=0.1,\alpha'=0.1$}  & \multicolumn{2}{c|}{$\alpha=0.2,\alpha'=0.2$} & \multicolumn{2}{c|}{$\alpha=0.2,\alpha'=0.2$} \\ 
         & \multicolumn{2}{c|}{$\beta=0,\beta'=0$}  
         & \multicolumn{2}{c|}{$\beta=0,\beta'=0$} 
         & \multicolumn{2}{c|}{$\beta=0,\beta'=0$} \\ 
         & \multicolumn{2}{c|}{$\gamma=0.9,\gamma'=0.9$}  & \multicolumn{2}{c|}{$\gamma=0.8,\gamma'=0.8$} & \multicolumn{2}{c|}{$\gamma=0.8,\gamma'=0.8$} \\ 
         \hline
         
&IoU  &F-score &IoU &F-score &IoU &F-score \\
\hline
\textbf{S1} &$76.69\pm0.989$ &$86.44\pm0.622$ &$74.08\pm1.812$ &$85.51\pm0.121$ &$77.93\pm0.537$ &$87.33\pm0.509$  \\
\hline
\textbf{S2} &$75.28\pm1.661$ &$85.18\pm1.060$ &$75.18\pm1.541$ &$85.90\pm0.782$ &$77.26\pm0.756$ &$87.13\pm0.381$  \\
\hline
\textbf{Ensemble} &$77.26\pm0.438$ &$87.49\pm0.593$ &$75.70\pm1.503$ &$86.02\pm0.974$ &$\mathbf{78.24\pm0.494}$ &$\mathbf{87.63\pm0.614}$ \\
\hline
 \hline
         
    \end{tabular}
    }
    
    \label{tab:41}
    \end{subtable}
% \vspace{2.5mm}
% **********************************************************************************
     \begin{subtable}{\linewidth}
     \centering
\resizebox{13cm}{!}{       
        \begin{tabular}{|c|c|c|c|c|c|c|}
        \hline
{KD} &\multicolumn{2}{c|} {V1} &\multicolumn{2}{c|} {V2} &\multicolumn{2}{c|} {V3} \\
        (off)   & \multicolumn{2}{c|}{$\alpha=0.2,\alpha'=0.2$}  & \multicolumn{2}{c|}{$\alpha=0.2,\alpha'=0.2$} & \multicolumn{2}{c|}{$\alpha=0.2,\alpha'=0.2$} \\ 
         & \multicolumn{2}{c|}{$\beta=0.8,\beta'=0.8$}  
         & \multicolumn{2}{c|}{$\beta=0.8,\beta'=0.8$} 
         & \multicolumn{2}{c|}{$\beta=0.8,\beta'=0.8$} \\ 
         & \multicolumn{2}{c|}{$\gamma=0,\gamma'=0$}  & \multicolumn{2}{c|}{$\gamma=0,\gamma'=0$} & \multicolumn{2}{c|}{$\gamma=0,\gamma'=0$} \\ 
         \hline
         
&IoU  &F-score &IoU &F-score &IoU &F-score \\
\hline
\textbf{T} &$76.86\pm0.791$ &$86.93\pm0.537$  &$76.86\pm0.791$ &$86.93\pm0.537$ &$76.86\pm0.791$ &$86.93\pm0.537$\\
\hline
\textbf{S1} &$75.81\pm0.643$ &$86.22\pm0.381$ &$75.45\pm0.410$ &$86.01\pm0.261$ &$76.81\pm1.432$ &$86.65\pm1.381$ \\
\hline
\textbf{S2} &$75.77\pm0.410$ & $86.11\pm0.956$&$76.29\pm1.576$ &$86.50\pm0.989$  &$77.59\pm0.931$ &$86.99\pm0.728$  \\
\hline
\textbf{Ensemble} &$77.28\pm0.356$ &$87.56\pm0.274$&$77.79\pm1.283$ &$87.29\pm1.576$  &$\mathbf{78.87\pm0.452}$ &$\mathbf{87.93\pm0.390}$ \\
\hline
 \hline
         
    \end{tabular}
    }
    
    \label{tab:42}
    \end{subtable}
% \vspace{2.5mm}
% **********************************************************************************
     \begin{subtable}{\linewidth}
     \centering
\resizebox{13cm}{!}{       
        \begin{tabular}{|c|c|c|c|c|c|c|}
        \hline
{KD} &\multicolumn{2}{c|} {V1} &\multicolumn{2}{c|} {V2} &\multicolumn{2}{c|} {V3} \\
         (on)& \multicolumn{2}{c|}{$\alpha=0.1,\alpha'=0.1$}  & \multicolumn{2}{c|}{$\alpha=0.2,\alpha'=0.2$} & \multicolumn{2}{c|}{$\alpha=0.1,\alpha'=0.2$} \\ 
         & \multicolumn{2}{c|}{$\beta=0.9,\beta'=0.9$}  
         & \multicolumn{2}{c|}{$\beta=0.8,\beta'=0.8$} 
         & \multicolumn{2}{c|}{$\beta=0.9,\beta'=0.8$} \\ 
         & \multicolumn{2}{c|}{$\gamma=0,\gamma'=0$}  & \multicolumn{2}{c|}{$\gamma=0,\gamma'=0$} & \multicolumn{2}{c|}{$\gamma=0,\gamma'=0$} \\ 
         \hline
         
&IoU  &F-score &IoU &F-score &IoU &F-score \\
\hline
\textbf{T} &$75.09\pm0.848$ &$85.11\pm0.558$  &$76.64\pm0.593$ &$86.77\pm0.381$ &$76.27\pm0.516$ &$86.71\pm0.190$\\ \hline
\textbf{S1} &$76.85\pm0.945$ &$86.91\pm0.601$ &$77.74\pm0.254$ &$87.47\pm0.162$ &$78.51\pm0.473$ &$88.09\pm0.452$ \\ \hline
\textbf{S2} &$77.52\pm0.466$ &$86.65\pm0.360$ &$77.51\pm0.367$ &$87.33+0.212$ &$78.04\pm0.226$ &$87.01\pm0.967$ \\ \hline
\textbf{Ensemble} &$78.76\pm0.565$ &$88.23\pm0.763$ &$78.41\pm0.141$ &$87.90\pm0.130$ &$\mathbf{79.23\pm0.494}$ &$\mathbf{88.37\pm0.322}$ \\
\hline
 \hline
         
    \end{tabular}
    }
    
    \label{tab:43}
    \end{subtable}
% \vspace{1mm}
% **********************************************************************************

     \begin{subtable}{\linewidth}
     \centering
\resizebox{13cm}{!}{       
        \begin{tabular}{|c|c|c|c|c|c|c|}
        \hline
\multirow{5}*{KD + ML} &\multicolumn{2}{c|} {V1} &\multicolumn{2}{c|} {V2} &\multicolumn{2}{c|} {V3} \\
         & \multicolumn{2}{c|}{$\alpha=0.2,\alpha'=0.2$}  & \multicolumn{2}{c|}{$\alpha=0.2,\alpha'=0.2$} & \multicolumn{2}{c|}{$\alpha=0.1,\alpha'=0.1$} \\ 
         & \multicolumn{2}{c|}{$\beta=0.4,\beta'=0.4$}  & \multicolumn{2}{c|}{$\beta=0.4,\beta'=0.4$} & \multicolumn{2}{c|}{$\beta=0.45,\beta'=0.45$} \\ 
         & \multicolumn{2}{c|}{$\gamma=0.4,\gamma'=0.4$}  & \multicolumn{2}{c|}{$\gamma=0.4,\gamma'=0.4$} & \multicolumn{2}{c|}{$\gamma=0.45,\gamma'=0.45$} \\ 
         \hline
         
&IoU  &F-score &IoU &F-score &IoU &F-score \\
\hline
\textbf{T} &$76.20\pm0.247$ &$86.08\pm0.749$ &$78.03\pm0.982$ &$87.70\pm0.685$  &$78.06\pm1.130$ &$87.61\pm1.330$\\ \hline
\textbf{S1} &$76.31\pm0.883$ &$86.56\pm0.565$ &$78.20\pm0.792$ &$87.76\pm0.459$ &$78.66\pm0.542$ &$88.46\pm0.506$  \\ \hline
\textbf{S2} &$77.00\pm0.707$ &$87.01\pm0.473$  &$78.94\pm1.440 $ &$88.23\pm0.905$ &$79.39\pm0.491$ &$88.84\pm0.367$ \\ \hline
\textbf{Ensemble} &$78.83\pm0.275$ &$88.09\pm0.070$  &$79.63\pm0.700$ &$88.31\pm0.638$ &\textcolor{blue}{\bm{$80.12\pm0.597$}} &\textcolor{blue}{\bm{$89.52\pm0.556$}}\\
\hline
 \hline
         
    \end{tabular}
    }
    
    \label{tab:44}
    \end{subtable}
% \vspace{2.5mm}

\label{tab:4}
\end{table*}

\section{Results and Discussion}

For a baseline comparison, we first evaluated the performance of standalone pre-trained ResNet50, ResNet18, and MobileNet models for both classification and segmentation tasks, as shown in Table \ref{tab:2}. Additional baseline performance metrics on the HAM10000 dataset are presented in Table \ref{tab:22} in the supplementary material.

\subsection{Classification Task}

The results of metastatic tissue classification using ResNet50-ResNet18 combination for teacher and student networks are shown in Table \ref{tab:3}. We observe that for each of the individual distillation techniques, the knowledge diversification paradigm (V3), where both predictions and features are shared, provides the best performance in terms of classification accuracy. Similarly, a comparison across different distillation techniques for a specific learning strategy shows that the combined KD + ML approach is superior, corroborating the findings of \cite{UNiyaz}. Expectedly, the combined KD + ML model trained with the knowledge diversification paradigm (V3) outperforms all other models. When compared with the conventional KD-only or ML-only models that share only predictions (V1), the proposed model provides an average improvement of $2\%$ in classification accuracy. Lastly, it can also be observed that in addition to the ensemble accuracy, the V3 learning strategy also improves the performance of individual student networks. 

{To demonstrate the generalizability of our proposed method across different datasets, we repeated the above experiments using the HAM10000 dataset for the lesion classification task. Results of this experiment, presented in Table {\ref{tab:99}} of the supplementary material, depict similar trends in the performance of various models.} Furthermore, to evaluate the robustness of our proposed approach across different architectures, the metastatic tissue classification experiments were repeated using the ResNet50-MobileNet combination for teacher and student networks. Results of these experiments are shown in Table \ref{tab:10} of the supplementary material. Notably, our proposed model demonstrated consistent improvement in performance across different network architectures and datasets, showcasing similar trends in the context of various distillation techniques, learning strategies, and their combinations.

 \begin{figure*}[t]
% \captionsetup{labelfont=bf}
% % \flushleft
% % \begin{flushleft}
% \hspace*{1.3cm} 
\centering
\includegraphics[width=0.8\linewidth]{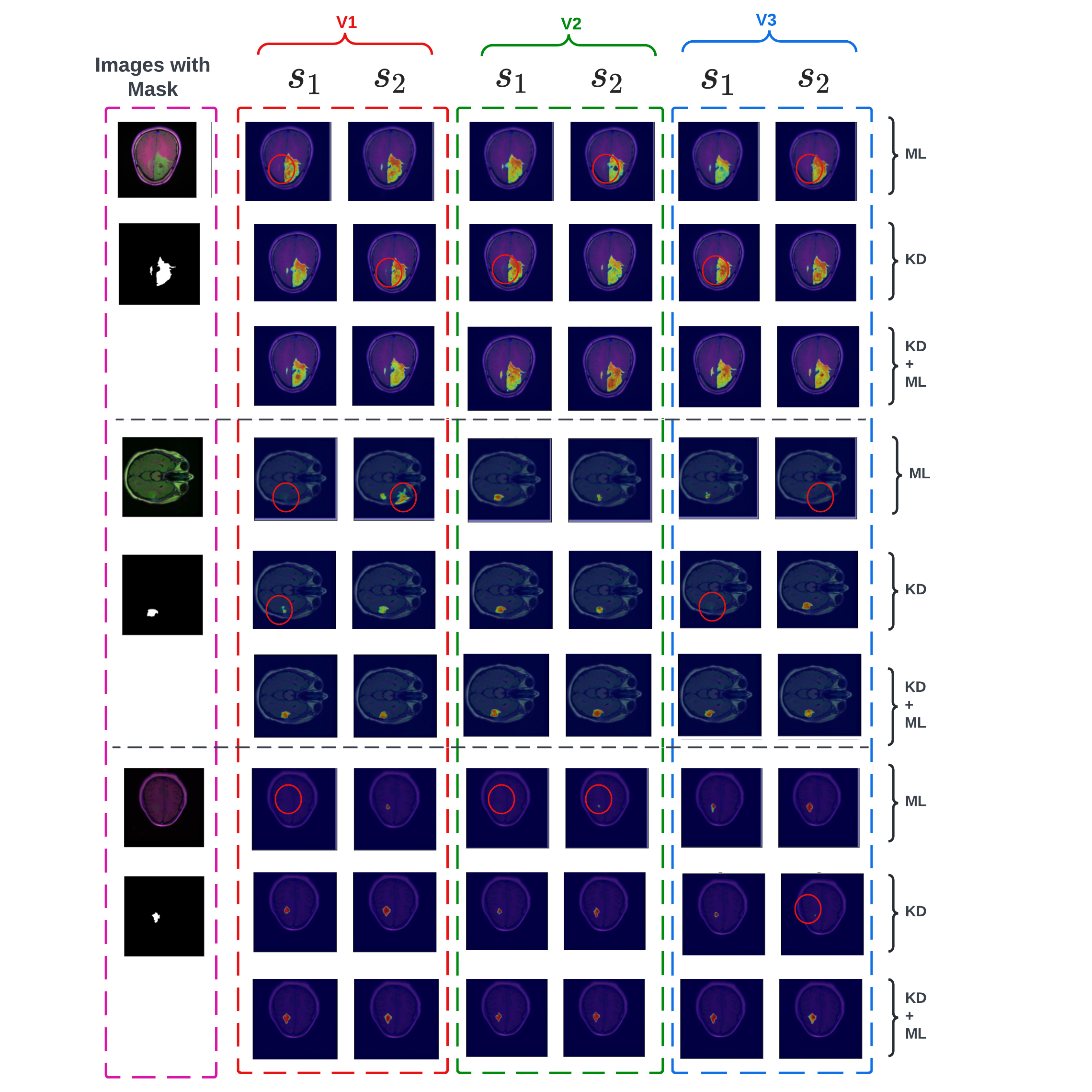}
\caption{Heatmap visualization of individual student predictions ($s_1$ and $s_2$) for three hard-to-segment examples. Predictions are shown for KD only, ML only, and KD + ML models trained with only V1 (predictions only), V2 (features only), and a diverse knowledge paradigm, V3 (both predictions and features), along with the original input image and the corresponding ground-truth mask. Red circles highlight regions where models failed to detect tumor segments.}
%\caption{Visualization to show the significance of the KD + ML approach using mixed information-sharing: The individual students suffer in segmenting the tumor in KD only and ML only for some hard-to-segment examples, whereas in our proposed approach, both the students perform better (The red circle denotes the missed regions). }
\label{fig:3}
% \end{flushleft}
\end{figure*}
\begin{figure*}[htbp]
% \captionsetup{labelfont=bf}
\centering
\includegraphics[width=0.8\linewidth]{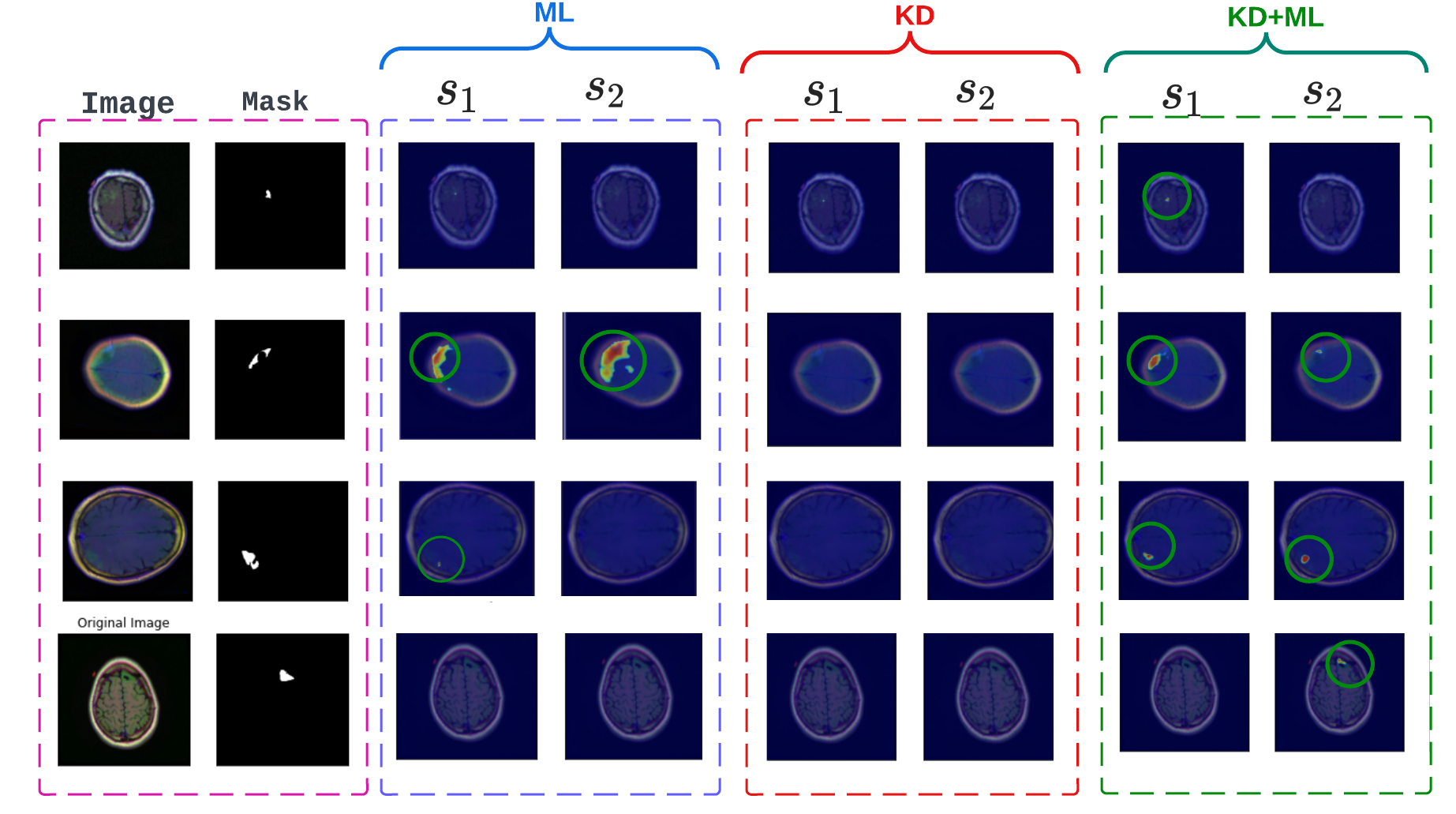}
\caption{Heatmap visualization of individual student predictions ($s_1$ and $s_2$) for four hard-to-segment examples. Predictions are shown for KD only, ML only, and KD + ML models trained using diverse knowledge paradigm V3 (both predictions and features) along with the original input image and corresponding ground-truth mask. Green circles highlight the tumor segments detected.}
%\caption{Visualization to depict the importance of combining KD with ML: Heatmaps of individual student predictions for some hard-to-segment examples with KD only, ML only, and KD + ML using mixed information-sharing V3 (predictions and features)(The green circle denotes the detected regions).}
\label{fig:4}
\end{figure*}
\subsection{Segmentation Task}

The corresponding results of U-Net with ResNet50-ResNet18 network architectures for the segmentation task are reported in Table \ref{tab:4}. These results depict similar trends observed for the classification task, further emphasizing the importance of combining KD with ML and the influence of knowledge diversification. Moreover, it establishes the generalizability of the proposed approach to more than one type of task.

To better appreciate the significance of the proposed approach, we provide a qualitative comparison of different distillation techniques and information-sharing strategies using individual student predictions for some hard-to-segment test samples. Figure \ref{fig:3} extensively compares all the models for three test samples. In general, it can be observed that KD-only and ML-only models struggle with the segmentation of small regions of interest. It can also be noticed that for conventional models, only one of the two students manages to predict these small regions. In our proposed approach, an ensemble of student predictions is used for the final predictions, facilitating the student networks to consistently predict the region of interest and yield optimal performance. Moreover, student networks trained using a diverse knowledge paradigm demonstrate a superior ability to discern finer structures compared to other models.

We demonstrate the importance of combining KD with ML by comparing all models trained with a knowledge diversification paradigm in Figure \ref{fig:4}. It can be noticed that the combined KD + ML model successfully detects these small regions of interest from these hard sample images where KD-only and ML-only models fail. Similarly, to underscore the significance of knowledge diversification over other learning strategies, we show sample predictions from the combined KD + ML model trained with V1 (predictions only), V2 (features only), and a diverse knowledge paradigm,  V3 (both predictions and features) (Figure \ref{fig:7} in the supplementary material) where we observe that the V3 strategy helps detect small and fine regions of interest better than V1 and V2.

{We repeated the above experiments using the HAM10000 dataset for the lesion segmentation task, and the corresponding results are presented in Table {\ref{tab:115}} of the supplementary material. Moreover, the LGG segmentation task was repeated using the ResNet50-MobileNet combination for the teacher-student networks in the U-Net. Consistent improvement in performance observed in these additional experiments further emphasizes the reliability of the KD + ML model with a knowledge diversification strategy.}

\subsection{Comparison with existing KD techniques}
Table \ref{tab:9} in the supplementary material compares our proposed approach with existing knowledge distillation methods. For the classification task, we conducted a comprehensive comparison against a range of knowledge distillation approaches, including Fitnet \cite{romeo}, AT \cite{at}, SP \cite{sp}, SRRL \cite{srrl}, VID \cite{vid}, SimKD \cite{SimKD}, and SemCKD \cite{semckd}. A consistent setup was maintained to ensure fairness, employing a  ResNet50 model as the teacher and ResNet18 as the student model for all existing methodologies.

In addition, for the segmentation task, we compared our approach with well-known segmentation methods like SKD \cite{skd}, IFVD \cite{ifvd}, CWD \cite{cwd}, and DSD \cite{dsd}, originally designed for computer vision semantic segmentation tasks. To provide a more comprehensive analysis, we also implemented AT \cite{at} and Fitnet \cite{romeo}, even though they were not initially designed for segmentation tasks. Additionally, we included a domain-specific segmentation approach, EMKD \cite{seg}. Throughout this evaluation, we maintained consistency by using a ResNet50 as the backbone of U-Net for the teacher model and ResNet18 for the student model.

This extensive comparative analysis offers valuable insights into the effectiveness and performance of our proposed method in relation to a diverse set of state-of-the-art knowledge distillation and segmentation techniques
\subsection{Explainability}

\begin{figure}
\centering
% \captionsetup{labelfont=bf}
\includegraphics[width=0.72\linewidth]{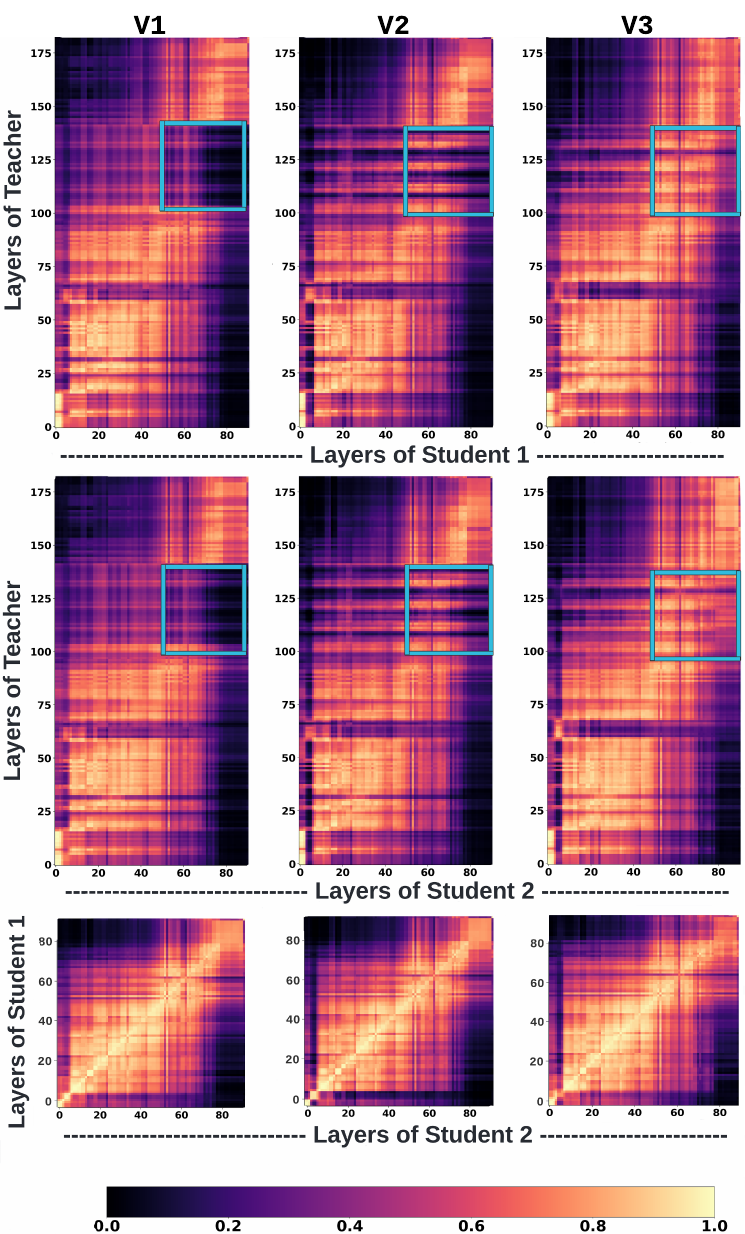}
\caption{CKA plots visualizing the similarity of learned representation between different layers of teacher and student networks using KD + ML (V3) model.}
\label{fig:5}
\end{figure}

To understand the effect of using different learning styles for the classification task, we used the popular Centered Kernel Alignment (CKA) metric \cite{cka} to measure and compare the similarity of learned representations of different layers of the teacher and student networks. As highlighted (blue box) in Figure \ref{fig:5}, the CKA plots show increased similarity between the higher layers of the teacher and final layers of the student networks for the knowledge diversification paradigm (V3). As the deeper layers of a network are considered task-specific, this increased similarity could potentially explain the improved performance of KD + ML with V3 compared to all other knowledge distillation and sharing strategies (Figure \ref{fig:8} and \ref{fig:9} in the supplementary material). Finally, Figure \ref{fig:6} shows the ensemble segmentation masks of different knowledge-sharing models trained with the knowledge diversification paradigm (V3). 

In addition,  fidelity is another important criterion for evaluating knowledge distillation models, where we expect the student predictions to match the teacher rather than achieve higher accuracy. We conducted some preliminary experiments to see how fidelity between students and teachers is affected by KD + ML as well as different learning styles. We observed that fidelity is not consistently improved for all combinations of experiments (Table \ref{tab:5} and \ref{tab:6} in the supplementary material). This is expected as the students are forced to not just match with the teacher but also with other students. Future efforts can be directed towards a combined increase in accuracy as well as fidelity of the student networks. 

\section{Conclusions}
\begin{figure*}
% \captionsetup{labelfont=bf}
\centering
\includegraphics[width=0.5\linewidth]{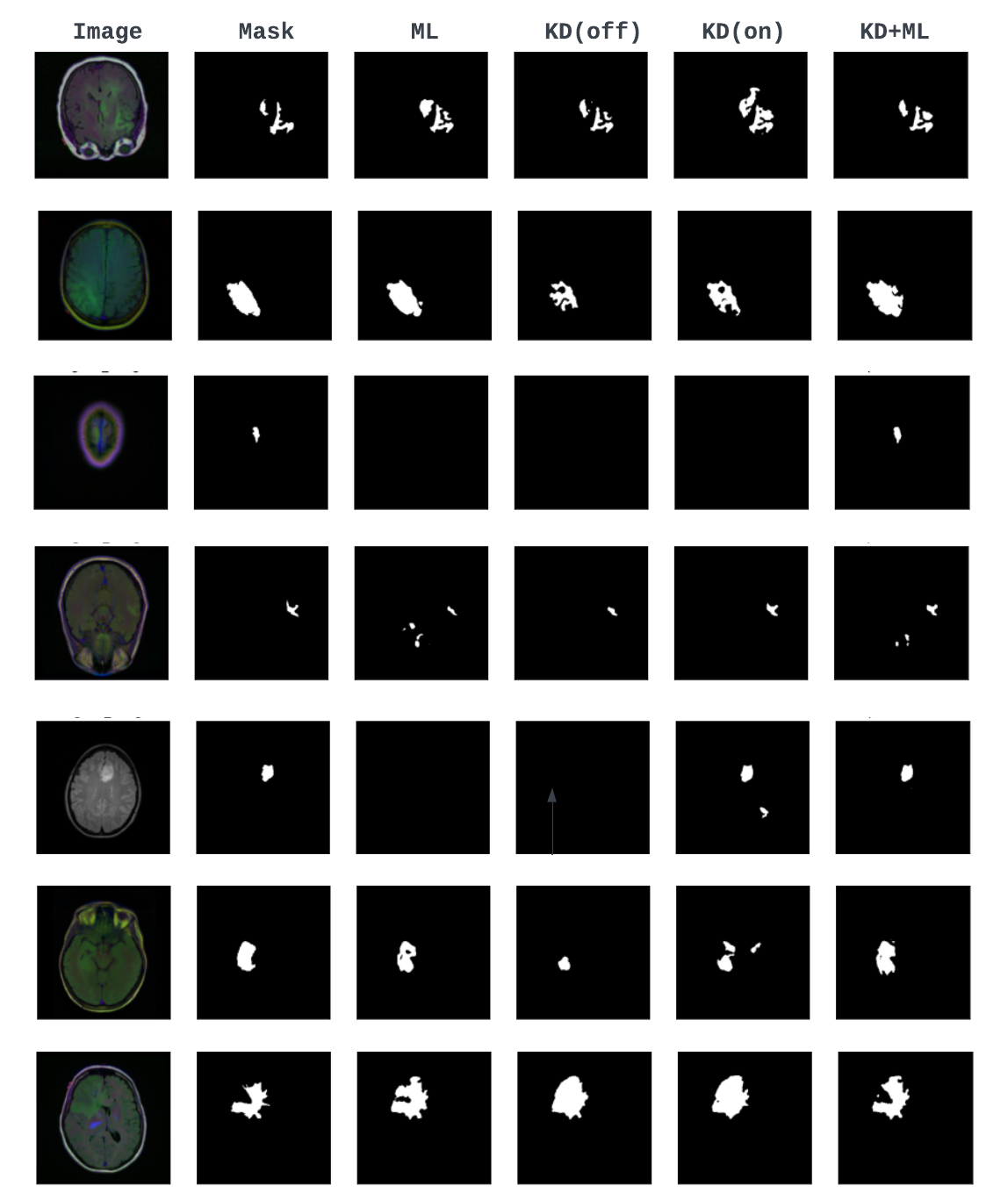}
\caption{Comparison of the quality of segmentations obtained using different knowledge-sharing models trained with a diverse knowledge paradigm, V3 (both predictions and features). The proposed approach generates segmentation masks that are much finer than other models and closer to the ground truth.}
\label{fig:6}
\end{figure*}
%In this work, we propose to enhance the conventional KD technique by combining it with ML in a single-teacher and two-student framework where students not only learn from the teacher but also from each other. Furthermore, knowledge diversification is achieved by training each student with different types of information extracted from the teacher and the other student. Extensive experiments on metastatic tissue classification, brain tumor segmentation, and dermatological classification and segmentation tasks conducted using four different distillation techniques and three different learning strategies demonstrate that our proposed approach outperforms existing knowledge distillation approaches. Specifically, our combined KD and ML model trained with knowledge diversification provides an average improvement of $2\%$ in classification accuracy over conventional KD or ML techniques (with similar network configuration) that only use predictions. The improvement in performance provided by the proposed approach is significant in the context of model compression. In practice, our proposed model can be adapted to a range of medical imaging tasks that necessitate lightweight networks for deployment in real-world scenarios without compromising performance.

In this work, we propose an enhanced knowledge transfer protocol that embraces the concept of different learning styles through knowledge diversification. Within the framework of a single-teacher, multi-student network that emulates classroom dynamics, the teacher network imparts knowledge to the student networks in various forms, such as predictions to one student and feature maps to another. Furthermore, we enhance individual student learning by fostering the exchange of diversified knowledge among students. Extensive experiments were conducted on metastatic tissue classification, brain tumor segmentation, and dermatological classification and segmentation tasks involving four different distillation techniques and three distinct learning strategies. These experiments showcase the superiority of our proposed approach over the existing knowledge distillation methods. In particular, our combined KD and ML model, trained with knowledge diversification, yields an average improvement of 2\% in classification accuracy compared to conventional KD or ML techniques with similar network configurations, which rely solely on predictions. The improvement in performance provided by the proposed approach is significant in the context of model compression. In practice, our proposed model can be adapted to a range of medical imaging tasks that necessitate lightweight networks for deployment in real-world scenarios without compromising performance.

\bibliography{reference_proj2}

\clearpage
\onecolumn
\setcounter{section}{0}
\renewcommand{\thesection}{\Alph{section}}
\section{Supplementary Material}

% %############################## 
% % Figures
% %##################################

\subsection{Teacher-Student information sharing}

%%%%%%%%%%%%%%%%%%%%%%%%%%%%%%%%%%%%%%%%%%%%%%%%%%%%%%%%%%%%%%%%%%%%%%%%%%%%%%%%%%%%%%

\begin{table}[!htp]\centering
\caption{A systematic representation of the different loss terms combinations that generate various combinations of distillation techniques and learning strategies.}

\begin{tabular}{ccccccccccc}\toprule
\multirow{2}{*}{Method} &\multirow{2}{*}{Version} &\multicolumn{2}{c}{$L_{KD}$} &\multicolumn{2}{c}{$L'_{KD}$} &\multicolumn{2}{c}{$L_{ML}$} &\multicolumn{2}{c}{$L'_{ML}$} \\\cmidrule{3-10}
& &$L_{KD_{p}}$ &$L_{KD_{f}}$ &$L'_{KD_{p}}$ &$L'_{KD_{f}}$ &$L_{ML_{p}}$ &$L_{ML_{f}}$ &$L'_{ML_{p}}$ &$L'_{ML_{f}}$ \\
\cmidrule{1-10}
\multirow{3}{*}{KD} &V1 &\checkmark & $\times$ &\checkmark &$\times$ &$\times$ &$\times$ &$\times$ &$\times$ \\
\cmidrule{2-10}
&V2 &$\times$ &\checkmark & $\times$ &\checkmark &$\times$ &$\times$ &$\times$ &$\times$ \\\cmidrule{2-10}
&V3 &\checkmark &$\times$ &$\times$ &\checkmark &$\times$ &$\times$ &$\times$ &$\times$ \\\cmidrule{1-10}
\multirow{3}{*}{ML} &V1 &$\times$ &$\times$ &$\times$ &$\times$ &\checkmark &$\times$ &\checkmark &$\times$ \\\cmidrule{2-10}
&V2 &$\times$ &$\times$ &$\times$ &$\times$ &$\times$ &\checkmark &$\times$ &\checkmark \\\cmidrule{2-10}
&V3 &$\times$ &$\times$ &$\times$ &$\times$ &\checkmark &$\times$ &$\times$ &\checkmark \\\cmidrule{1-10}
\multirow{3}{*}{KD + ML} &V1 &\checkmark &$\times$ &\checkmark &$\times$ &\checkmark &$\times$ &\checkmark &$\times$ \\\cmidrule{2-10}
&V2 &$\times$ &\checkmark &$\times$ &\checkmark &$\times$ &\checkmark &$\times$ &\checkmark \\\cmidrule{2-10}
&V3 &$\times$ &\checkmark &\checkmark &$\times$ &\checkmark &$\times$ &$\times$ &\checkmark \\\midrule
\bottomrule
\end{tabular}
\label{tab:13}
\end{table}

\begin{figure*}[htbp]
\centering
\includegraphics[width=13cm, height=8cm]{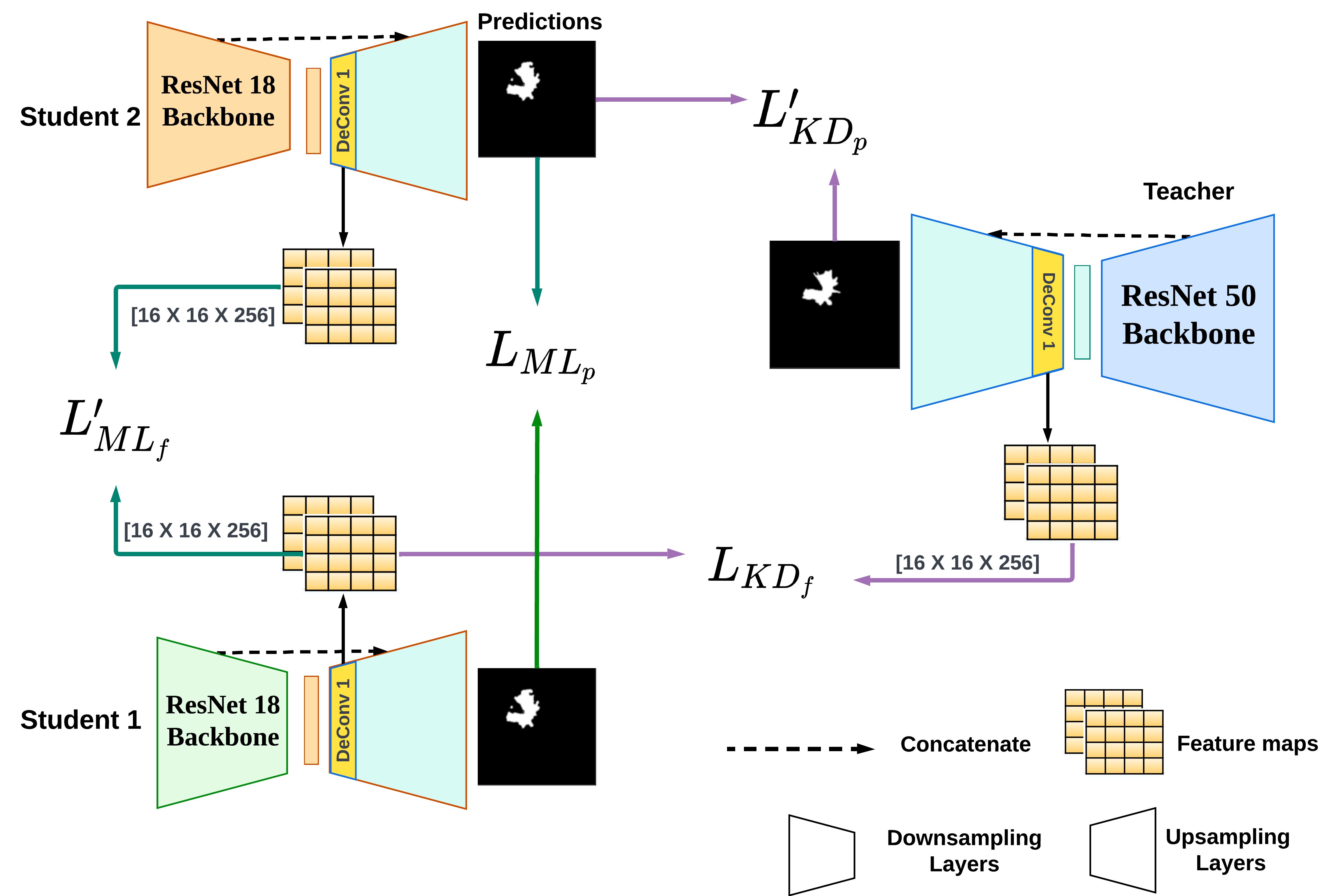}
\caption{{Overview of our proposed model that combines Knowledge Distillation (KD) with Mutual Learning (ML) in a single-teacher, two-student framework, leveraging a diversified knowledge-sharing strategy in the segmentation task. The loss terms $L$ and $L'$ represent the respective losses for each student, while $L'_{KD_{p}}$ and $L_{KD_{f}}$ capture the knowledge distillation losses between the teacher and concerned student over predictions and features, respectively. Similarly, $L_{ML_{p}}$ and $L'_{ML_{f}}$ denote the mutual learning losses between the two students over predictions and features, respectively. Here ResNet50 is used as a teacher network and ResNet18 as a student network  }}
\label{fig:11}
\end{figure*}

{
\subsection{Teacher-Student network details}
\textbf{ResNet 50 as a teacher model:} ResNet-50, a deep convolutional neural network (CNN) architecture, was introduced in 2015 \cite{resnet50} and has been highly influential in the field of image classification since then. It incorporates a unique concept called residual learning, which employs skip connections to enable learning residual functions instead of directly fitting the desired mapping. The model is pretrained on the ImageNet dataset, benefiting from pre-initialized weights, rendering it a powerful architecture for image analysis tasks. The decision to use ResNet-50 as the teacher model is driven by its exceptional ability to capture complex features.

\textbf{Resnet18 as a student model: }  ResNet-18 serves as a suitable student model in our approach. It is a lighter variant within the ResNet architecture, with fewer layers. Despite its reduced complexity, ResNet-18 demonstrates commendable performance in classification tasks. Similar to ResNet-50, we initialize the model weights using pre-trained weights from the ImageNet dataset \cite{classificationmodels}, enabling effective knowledge transfer. The architecture details for both ResNet50 and ResNet18 are defined in Table \ref{tab:7}.  

\textbf{MobileNet as a student model: }
MobileNet \cite{howard2017mobilenets} is a family of lightweight deep neural network architectures tailored for mobile and embedded vision applications. It utilizes depthwise separable convolutions, employing separate convolutional filters for each input channel followed by a 1x1 pointwise convolution to reduce parameters and computation significantly. This approach achieves efficient inference on resource-constrained devices while maintaining reasonable accuracy. 

\textbf{U-Net with ResNet-50/ResNet-18 as Backbone:} By
leveraging pre-trained ResNet-50/ResNet-18 as the foundational architecture for U-Net\cite{unet}, we harness their robust feature
representation capabilities to enhance the accuracy of segmentation tasks significantly. The strategic use of skip
connections within the U-Net framework plays a crucial
role in preserving intricate details and spatial information
during upsampling, leading to superior segmentation performance. U-Net augmented with ResNet-50/ResNet-18
backbones prove to be a versatile and effective choice for
diverse image segmentation applications, including medical, semantic, and instance segmentation. Our methodology adopts U-Net with ResNet-50 as the teacher model,
encompassing around 33 million parameters, and
U-Net with ResNet-18 as the student network, comprising
around 14 million parameters. The architecture details for U-Net are defined in Table \ref{tab:8}.  

\textbf{U-Net with MobileNet as backbone:}
Integrating pre-trained MobileNet as the backbone of U-Net, the architecture leverages MobileNet's computational performance while utilizing U-Net's superior segmentation capabilities. This combination aims to achieve efficient image segmentation, making it well-suited for applications in resource-constrained environments or real-time processing scenarios. In this study, U-Net with MobileNet as the student network, comprising around 8 million parameters.
}

% %############################## 
% % Classification architecture
% %##################################

\begin{table}[htbp]
\centering
\caption{Architecture of ResNet-50 and Resnet-18 network }
\resizebox{9cm}{!}{ 
\begin{tabular}{|c|c|c|c|} 
\toprule

layer name & Output Shape & 18-Layer & 50-Layer\\
\midrule
{Conv1} & $112 \times 112$ &  \multicolumn{2}{c|}{$7 \times 7,64$, stride 2 } \\
\hline
& &  \multicolumn{2}{c|}{$3 \times 3$,maxpool, stride 2 }  \\
\cmidrule{3-4}
 % \vspace{4pt}
\multirow{2}{*}{Conv2.x} & \multirow{2}{*}
{$56 \times 56$}  & $\left[\begin{array}{c}3 \times 3,64 \\ 3 \times 3,64\end{array}\right] \times 2$ & $\left[\begin{array}{c}1 \times 1,64 \\ 3 \times 3,64 \\ 1 \times 1,256\end{array}\right] \times 3$ \\
\midrule

Conv3.x & $28 \times 28$  & $\left[\begin{array}{c}3 \times 3,128 \\ 3 \times 3,128\end{array}\right] \times 2$ & $\left[\begin{array}{c}1 \times 1,128 \\ 3 \times 3,128 \\ 1 \times 1,512\end{array}\right] \times 3$ \\
\midrule
Conv4.x & $14 \times 14$  & $\left[\begin{array}{c}3 \times 3,256 \\ 3 \times 3,256\end{array}\right] \times 2$ & $\left[\begin{array}{c}1 \times 1,256 \\ 3 \times 3,256 \\ 1 \times 1,1024\end{array}\right] \times 6$  \\
\midrule
Conv5.x & $7 \times 7$  & $\left[\begin{array}{c}3 \times 3,512 \\ 3 \times 3,512\end{array}\right] \times 2$ & $\left[\begin{array}{c}1 \times 1,512 \\ 3 \times 3,512 \\ 1 \times 1,2048\end{array}\right] \times 3$  \\
\midrule
Conv6 & $7 \times 7$  & - & $\left[\begin{array}{c}1 \times 1,512 \\ \end{array}\right] \times 1$  \\
\midrule
\end{tabular}
}
\label{tab:7}
\end{table}

\begin{table*}[htbp]
\centering
\caption{Architecture of U-Net with ResNet-50 and Resnet-18 as a backbone}
\resizebox{13cm}{!}{ 
\begin{tabular}{|c|c|c|c|c|c|c|} 
\toprule
\multicolumn{4}{c|}{Downsampling Layers}&  \multicolumn{3}{|c|}{Upsampling Layers}\\
\toprule
layer name & Output Shape & 18-Layer & 50-Layer&Layer name & Output Shape & Upsample Block\\
\midrule
Conv1 & $128 \times 128$ &  \multicolumn{2}{c|}{$3 \times 3,64$, stride 2 } & DeConv 1 & $16 \times 16$  & $\left[\begin{array}{c}3 \times 3,256 \\ 3 \times 3,256\end{array}\right] \times 2$ \\
\midrule
 % \vspace{4pt}
Conv2.x & $64 \times 64$  & $\left[\begin{array}{c}3 \times 3,64 \\ 3 \times 3,64\end{array}\right] \times 2$ & $\left[\begin{array}{c}1 \times 1,64 \\ 3 \times 3,64 \\ 1 \times 1,256\end{array}\right] \times 3$ & DeConv 2 & $32 \times 32$  & $\left[\begin{array}{c}3 \times 3,128 \\ 3 \times 3,128\end{array}\right] \times 2$\\
\midrule

Conv3.x & $32 \times 32$  & $\left[\begin{array}{c}3 \times 3,128 \\ 3 \times 3,128\end{array}\right] \times 2$ & $\left[\begin{array}{c}1 \times 1,128 \\ 3 \times 3,128 \\ 1 \times 1,512\end{array}\right] \times 3$ & DeConv 3 & $64 \times 64$  & $\left[\begin{array}{c}3 \times 3,64 \\ 3 \times 3,64\end{array}\right] \times 2$\\
\midrule
Conv4.x & $16 \times 16$  & $\left[\begin{array}{c}3 \times 3,256 \\ 3 \times 3,256\end{array}\right] \times 2$ & $\left[\begin{array}{c}1 \times 1,256 \\ 3 \times 3,256 \\ 1 \times 1,1024\end{array}\right] \times 6$ &
DeConv 4 & $128 \times 128$  & $\left[\begin{array}{c}3 \times 3,32 \\ 3 \times 3,32\end{array}\right] \times 2$ \\
\midrule
Conv5.x & $8 \times 8$  & $\left[\begin{array}{c}3 \times 3,512 \\ 3 \times 3,512\end{array}\right] \times 2$ & $\left[\begin{array}{c}1 \times 1,512 \\ 3 \times 3,512 \\ 1 \times 1,2048\end{array}\right] \times 3$ & DeConv 5 & $256 \times 256$  & $\left[\begin{array}{c}3 \times 3,16 \\ 3 \times 3,16\end{array}\right] \times 2$ \\
\midrule
\end{tabular}}
\label{tab:8}
\end{table*}

\newpage
\subsection{Results:}
%############################## 
% classification table MobileNet
%##################################

\begin{table*}[htbp]
% \captionsetup{labelfont=bf}
\centering
% \captionsetup{font=small}
      \caption{Performance comparison of classification accuracy for Histopathologic Cancer Detection dataset using four different distillation techniques: ML - Mutual Learning, KD (on) - online Knowledge Distillation, KD (off) - offline Knowledge Distillation, and KD + ML – combined KD \& ML; and three different learning strategies -  V1 (predictions only), V2 (features only) and a diverse knowledge paradigm, V3 (both predictions and features). Here, ResNet50 is used as the teacher network and  MobileNet as the student network }
    \begin{subtable}{\linewidth}
    \centering
  \resizebox{10.3cm}{!}{       
        \begin{tabular}{|c|c|c|c|}
        \hline
         \multirow{5}*{ML}&{V1} & V2 & V3   \\
         & ($\alpha=0.2,\alpha'=0.2$)   & ($\alpha=0.1,\alpha'=0.2$) & (${\alpha=0.1},\alpha'=0.2$)\\
         & ($\beta=0,\beta'=0$)   & ($\beta=0,\beta'=0$)& (${\beta=0},\beta'=0$)  \\
         & ($\gamma=0.8,\gamma'=0.8$)   & ($\gamma=0.9,\gamma'=0.8$) & (${\gamma=0.9},\gamma'=0.8$)\\
         \hline      
         
         \textbf{S1} &$92.67\pm0.268$ &$91.95\pm0.523$ &$92.21\pm0.410$  \\
 \hline
\textbf{S2} &$92.83\pm0.142$ &$91.69\pm0.654$  &$92.99\pm0.325$\\
 \hline
\textbf{Ensemble} &$93.43\pm0.170$ &$92.18\pm0.675$ &$\mathbf{93.53\pm0.127} $ \\
         
\hline
 \hline
         
    \end{tabular}
    }
    
    \label{tab:101}
    \end{subtable}
% \vspace{2.5mm}
    \begin{subtable}{\linewidth}
    \centering
\resizebox{10.2cm}{!}{ 
            \begin{tabular}{|c|c|c|c|}
        \hline
       \multirow{5}*{KD(off)}&{V1} & V2 & V3   \\
         
         & ($\alpha=0.2,\alpha'=0.2$)   & ($\alpha=0.2,\alpha'=0.2$) & (${\alpha=0.2},\alpha'=0.2$)\\
         & ($\beta=0.8,\beta'=0.8$)   & ($\beta=0.8,\beta'=0.8$)& (${\beta=0.8},\beta'=0.8$)  \\
         & ($\gamma=0,\gamma'=0$)   & ($\gamma=0,\gamma'=0$) & (${\gamma=0},\gamma'=0$)\\
         \hline
        \textbf{T} &$94.35\pm 0.763$  &$94.35\pm0.763$ &$94.35\pm0.763$\\
         \hline
\textbf{S1} & $93.39\pm0.070$ &$92.73\pm0.664$ &$93.61\pm0.353$ \\
         \hline
\textbf{S2} &$93.73\pm0.219$  &$93.72\pm0.339$ &$93.33\pm0.155$\\
\hline
\textbf{Ensemble} &$94.02\pm0.091$  &$93.97\pm0.120$ &$\mathbf{94.23\pm0.155}$\\
\hline
 \hline
         
    \end{tabular}
    }
    \label{tab:102}
    \end{subtable}{}

    % \vspace{2.5mm}
    \begin{subtable}{\linewidth}
    \centering
\resizebox{10.2cm}{!}{ 
            \begin{tabular}{|c|c|c|c|}
        \hline
 \multirow{5}*{KD(on)}&{V1} & V2 & V3   \\
          & ($\alpha=0.1,\alpha'=0.2$)   & ($\alpha=0.2,\alpha'=0.2$) & (${\alpha=0.2},\alpha'=0.2$)\\
         & ($\beta=0.9,\beta'=0.8$)   & ($\beta=0.8,\beta'=0.8$)& (${\beta=0.8},\beta'=0.8$)  \\
         & ($\gamma=0,\gamma'=0$)   & ($\gamma=0,\gamma'=0$) & (${\gamma=0},\gamma'=0$)\\
         \hline
         
         \hline
         \textbf{T} &$94.52\pm0.128$  &$94.45\pm0.268$ &$94.43\pm0.212$ \\
\hline
\textbf{S1} &$94.07\pm0.091$  &$94.02\pm0.070$ &$93.89\pm0.466$ \\
\hline
\textbf{S2} &$94.1\pm0.226$  &$93.35\pm0.170$ &$94.15\pm0.079$ \\
\hline
\textbf{Ensemble} &$94.12\pm0.028$ &$94.04\pm0.012$ &$\mathbf{94.47\pm0.593}$ \\

 \hline
 \hline
         
    \end{tabular}
   } 
       \label{tab:103}
    \end{subtable}{}

    % \vspace{2.5mm}
    \begin{subtable}{\linewidth}
    \centering
\resizebox{10.2cm}{!}{ 
            \begin{tabular}{|c|c|c|c|}
        \hline
         \multirow{5}*{KD + ML}&{V1} & V2 & V3   \\
         & ($\alpha=0.1,\alpha'=0.2$)   & ($\alpha=0.2,\alpha'=0.2$) & (${\alpha=0.2},\alpha'=0.4$)\\
         & ($\beta=0.45,\beta'=0.4$)   & ($\beta=0.4,\beta'=0.4$)& (${\beta=0.4},\beta'=0.3$)  \\
         & ($\gamma=0.45,\gamma'=0.4$)   & ($\beta=0.4,\beta'=0.4$) & (${\gamma=0.4},\gamma'=0.3$)\\
         
         \hline
\textbf{T} &$94.68\pm0.311$  &$94.72\pm0.213$ &$95.05\pm0.005$ \\
\hline
\textbf{S1} &$94.33\pm0.127$ &$93.81\pm0.270$ &$95.03\pm0.070$ \\
\hline
\textbf{S2} &$94.16\pm0.184$ &$94.04\pm0.512$ &$94.29\pm0.342$  \\
\hline
\textbf{Ensemble} &$94.95\pm0.077 $&$94.74\pm0.121 $ &$\textcolor{blue}{\bm{95.61\pm0.174}}$ \\
\hline

 \hline
         
    \end{tabular}
    }
       \label{tab:104}
    \end{subtable}{}

\label{tab:10}
\end{table*}

%############################## 
% Segmentation table Mobilenet
%##################################
\begin{table*}[htbp]
% \captionsetup{labelfont=bf}
\renewcommand{\arraystretch}{1.2}
% \captionsetup{font=small}
\centering
        \caption{Performance comparison for segmentation task using LGG dataset with IoU and F-score metrics using four different distillation techniques: ML - Mutual Learning, KD (on) - online Knowledge Distillation, KD (off) - offline Knowledge Distillation, and KD + ML – combined KD \& ML; and three different learning strategies -  (V1 (predictions only), V2 (features only) and a diverse knowledge paradigm, V3 (both predictions and features). Here, ResNet50 is used as the teacher network and  MobileNet as the student network}
    \begin{subtable}{\linewidth}
    \centering
    	% \hspace*{-100cm} 
\resizebox{13cm}{!}{       
        \begin{tabular}{|c|c|c|c|c|c|c|}
        \hline
\multirow{5}*{ML} &\multicolumn{2}{c|} {V1} &\multicolumn{2}{c|} {V2} &\multicolumn{2}{c|} {V3} \\
         & \multicolumn{2}{c|}{$\alpha=0.1,\alpha'=0.1$}  & \multicolumn{2}{c|}{$\alpha=0.2,\alpha'=0.2$} & \multicolumn{2}{c|}{$\alpha=0.1,\alpha'=0.2$} \\ 
         & \multicolumn{2}{c|}{$\beta=0,\beta'=0$}  
         & \multicolumn{2}{c|}{$\beta=0,\beta'=0$} 
         & \multicolumn{2}{c|}{$\beta=0,\beta'=0$} \\ 
         & \multicolumn{2}{c|}{$\gamma=0.9,\gamma'=0.9$}  & \multicolumn{2}{c|}{$\gamma=0.8,\gamma'=0.8$} & \multicolumn{2}{c|}{$\gamma=0.9,\gamma'=0.8$} \\ 
         \hline
         
&IoU  &F-score &IoU &F-score &IoU &F-score \\
\hline
\textbf{S1} &$70.35\pm0.655$ &$81.90\pm0.601$  &$69.65\pm0.315$ &$81.90\pm0.671$ &$70.76\pm0.445$ &$82.87\pm0.304$\\
\hline
\textbf{S2} &$69.65\pm0.902$ &$82.10\pm0.322$ &$68.55\pm1.180$ &$80.11\pm0.234$ &$69.81\pm1.302$ &$82.97\pm1.005$ \\
\hline

\textbf{Ensemble} &$71.48\pm0.484$ &$83.11\pm1.048$ &$70.14\pm1.712$ &$82.15\pm0.987$ &$\mathbf{73.05\pm0.894}$ &$\mathbf{83.95\pm0.903}$ \\
\hline
 \hline
         
    \end{tabular}
    }
    
    \label{tab:111}
    \end{subtable}
% \vspace{2.5mm}
% **********************************************************************************
     \begin{subtable}{\linewidth}
     \centering
\resizebox{13cm}{!}{       
        \begin{tabular}{|c|c|c|c|c|c|c|}
        \hline
{KD} &\multicolumn{2}{c|} {V1} &\multicolumn{2}{c|} {V2} &\multicolumn{2}{c|} {V3} \\
        (off)   & \multicolumn{2}{c|}{$\alpha=0.2,\alpha'=0.2$}  & \multicolumn{2}{c|}{$\alpha=0.2,\alpha'=0.2$} & \multicolumn{2}{c|}{$\alpha=0.2,\alpha'=0.2$} \\ 
         & \multicolumn{2}{c|}{$\beta=0.8,\beta'=0.8$}  
         & \multicolumn{2}{c|}{$\beta=0.8,\beta'=0.8$} 
         & \multicolumn{2}{c|}{$\beta=0.8,\beta'=0.8$} \\ 
         & \multicolumn{2}{c|}{$\gamma=0,\gamma'=0$}  & \multicolumn{2}{c|}{$\gamma=0,\gamma'=0$} & \multicolumn{2}{c|}{$\gamma=0,\gamma'=0$} \\ 
         \hline
         
&IoU  &F-score &IoU &F-score &IoU &F-score \\
\hline
\textbf{T} &$76.86\pm0.691$ &$86.93\pm0.537$  &$76.86\pm0.691$ &$86.93\pm0.537$ &$76.86\pm0.791$ &$86.93\pm0.537$\\
\hline

\textbf{S1} &$70.53\pm1.025$ &$81.83\pm0.622$ &$70.09\pm0.247$ &$81.43\pm0.169$ &$71.51\pm0.647$ &$82.74\pm0.169$  \\
\hline
\textbf{S2} &$70.29\pm0.994$ &$82.47\pm1.036$ &$70.18\pm0.247$ &$82.06\pm0.882$ &$72.09\pm0.329$ &$81.66\pm0.0.438$  \\
\hline
\textbf{Ensemble} &$71.51\pm0.477$ &$83.38\pm0.311$&$71.09\pm0.869$ &$82.49\pm1.586$  &$\mathbf{73.19\pm0.566}$ &$\mathbf{84.57\pm0.718}$ \\
\hline
 \hline
         
    \end{tabular}
    }
    
    \label{tab:112}
    \end{subtable}
% \vspace{2.5mm}
% **********************************************************************************
     \begin{subtable}{\linewidth}
     \centering
\resizebox{13cm}{!}{       
        \begin{tabular}{|c|c|c|c|c|c|c|}
        \hline
{KD} &\multicolumn{2}{c|} {V1} &\multicolumn{2}{c|} {V2} &\multicolumn{2}{c|} {V3} \\
         (on)& \multicolumn{2}{c|}{$\alpha=0.1,\alpha'=0.1$}  & \multicolumn{2}{c|}{$\alpha=0.2,\alpha'=0.2$} & \multicolumn{2}{c|}{$\alpha=0.1,\alpha'=0.2$} \\ 
         & \multicolumn{2}{c|}{$\beta=0.9,\beta'=0.9$}  
         & \multicolumn{2}{c|}{$\beta=0.8,\beta'=0.8$} 
         & \multicolumn{2}{c|}{$\beta=0.9,\beta'=0.8$} \\ 
         & \multicolumn{2}{c|}{$\gamma=0,\gamma'=0$}  & \multicolumn{2}{c|}{$\gamma=0,\gamma'=0$} & \multicolumn{2}{c|}{$\gamma=0,\gamma'=0$} \\ 
         \hline
         
&IoU  &F-score &IoU &F-score &IoU &F-score \\
\hline
\textbf{T} &$76.43\pm0.742$ &$86.64\pm0.473$  &$76.11\pm0.325$ &$86.07\pm0.205$ &$76.57\pm0.606$ &$86.07\pm0.887$\\ \hline
\textbf{S1} &$70.41\pm0.753$ &$81.93\pm0.223$ &$70.06\pm0.304$ &$80.96\pm0.856$ &$71.49\pm0.767$ &$83.27\pm0.163$ \\ \hline
\textbf{S2} &$70.66\pm0.435$ &$82.25\pm0.260$ &$70.67\pm0.441$ &$81.09+0.570$ &$71.77\pm0.876$ &$83.56\pm0.537$ \\ \hline
\textbf{Ensemble} &$72.76\pm0.491$ &$84.23\pm0.782$ &$71.41\pm0.366$ &$81.31\pm0.528$ &$\mathbf{73.92\pm0.622}$ &$\mathbf{85.03\pm0.410}$ \\
\hline
 \hline
         
    \end{tabular}
    }
    
    \label{tab:113}
    \end{subtable}
% \vspace{1mm}
% **********************************************************************************

     \begin{subtable}{\linewidth}
     \centering
\resizebox{13cm}{!}{       
        \begin{tabular}{|c|c|c|c|c|c|c|}
        \hline
\multirow{5}*{KD + ML} &\multicolumn{2}{c|} {V1} &\multicolumn{2}{c|} {V2} &\multicolumn{2}{c|} {V3} \\
         & \multicolumn{2}{c|}{$\alpha=0.2,\alpha'=0.2$}  & \multicolumn{2}{c|}{$\alpha=0.2,\alpha'=0.2$} & \multicolumn{2}{c|}{$\alpha=0.1,\alpha'=0.1$} \\ 
         & \multicolumn{2}{c|}{$\beta=0.4,\beta'=0.4$}  & \multicolumn{2}{c|}{$\beta=0.4,\beta'=0.4$} & \multicolumn{2}{c|}{$\beta=0.45,\beta'=0.45$} \\ 
         & \multicolumn{2}{c|}{$\gamma=0.4,\gamma'=0.4$}  & \multicolumn{2}{c|}{$\gamma=0.4,\gamma'=0.4$} & \multicolumn{2}{c|}{$\gamma=0.45,\gamma'=0.45$} \\ 
         \hline
         
&IoU  &F-score &IoU &F-score &IoU &F-score \\
\hline
\textbf{T} &$76.92\pm0.223$ &$86.71\pm0.735$ &$76.25\pm0.911$ &$87.89\pm0.786$  &$76.40\pm0.374$ &$87.81\pm0.240$\\ \hline
\textbf{S1} &$71.77\pm0.732$ &$82.64\pm1.106$ &$70.81\pm0.417$ &$82.31\pm0.642$ &$72.13\pm0.657$ &$83.76\pm0.526$  \\ \hline
\textbf{S2} &$71.53\pm1.601$ &$81.61\pm0.756$  &$71.38\pm0.791 $ &$81.92\pm0.551$ &$71.59\pm1.031$ &$82.94\pm0.767$ \\ \hline
\textbf{Ensemble} &$73.14\pm0.047$ &$84.48\pm0.070$  &$72.12\pm0.110$ &$83.53\pm0.247$ &\textcolor{blue}{\bm{$74.31\pm0.507$}} &\textcolor{blue}{\bm{$85.56\pm0.332$}}\\
\hline
 \hline
         
    \end{tabular}
    }
    
    \label{tab:114}
    \end{subtable}
% \vspace{2.5mm}

\label{tab:11}
\end{table*}

% %############################## 
% % Figures
% %##################################
\begin{figure}[htbp]
\centering
\captionsetup{labelfont=bf}
\includegraphics[width=0.8\linewidth]{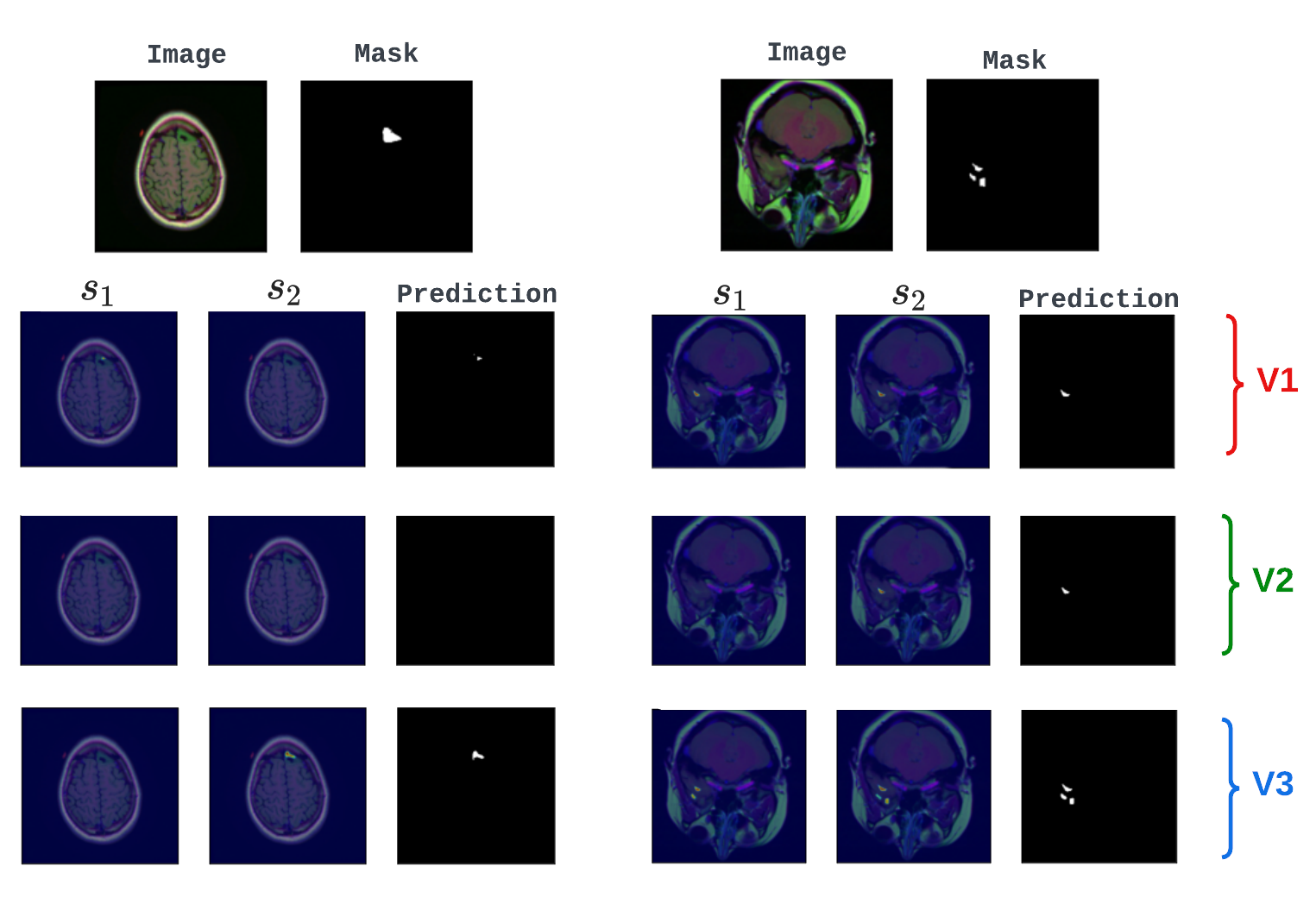}
\caption{Heatmaps of individual student predictions for some hard-to-segment examples with KD + ML model trained using V1 (predictions only), V2 (features only) and a diverse knowledge paradigm, V3 (both predictions and features). The heatmaps highlight the significance of integrating both predictions and features in the combined model (V3) over predictions only (V1) and feature only (V2)}
\label{fig:7}
\end{figure}

\newpage
\subsection{Comparison with existing KD techniques}
% %############################## 
% % Comparison table 
% %##################################
\begin{table}[htbp]
% \captionsetup{labelfont=bf}
\centering
% \captionsetup{font=small}
\caption{Comparison with other Knowledge Distillation methods in both Classification and Segmentation tasks. We used the ResNet50 and ResNet18 as the backbone architectures for teacher and student network respectively }\label{tab: }
\resizebox{14cm}{!}{ 
\begin{tabular}{|c|c||c|c|c|c|}\hline
\multicolumn{2}{|c||}{Classification} &\multicolumn{3}{c|}{Segmentation} \\\hline
Method &Accuracy &Method &IoU &F-score \\\hline
Fitnet\cite{romeo} &$94.78\pm0.197$ &Fitnet\cite{romeo} & $75.24\pm 0.261$ & $86.09\pm0.141$\\\hline
SRRL\cite{srrl} &$91.70\pm1.074$ &AT\cite{at} &$75.71\pm0.197$ & $86.18\pm0.134 $\\\hline
AT\cite{at} &$94.99\pm 0.01$4 &SKD\cite{skd} &$77.17\pm0.254$ &$87.51\pm0.494 $\\\hline
SP\cite{sp} &$94.56\pm0.311$ &IFVD\cite{ifvd} &$78.48\pm0.325$ & $88.58\pm0.183$ \\\hline
VID\cite{vid} &$95.34\pm0.374$ &CWD\cite{cwd} &$75.87\pm 0.657$ & $86.58\pm0.862$ \\\hline
SimKD\cite{SimKD} &$84.01\pm0.212$ &DSD\cite{dsd} &$77.47\pm0.084$ & $87.62\pm0.494$ \\\hline
SemCKD\cite{semckd} &$95.41\pm0.204$&EMKD\cite{seg} &$73.62\pm0.656$ & $84.29\pm1.576$ \\\hline
\textbf{Ours} &$\bm{96.68\pm0.584}$ & \textbf{Ours} & $\bm{80.12\pm0.597}$ & $\bm{89.52\pm0.556}$ \\\hline
\hline
\end{tabular}
}
\label{tab:9}
\end{table}

\subsection{Explainability}
%%%%%%%%%%%%%%%%%%%%%%%%%%%%%%%%%%%%%%%%%%%%%%%%%%%%%%%%%%%%%%%%%%%%%%%%%%%%%%%%%%%%%%

\begin{table}[htbp]
% \captionsetup{labelfont=bf}
\centering
% \captionsetup{font=small}
      \caption{Fidelity comparison of ResNet50-ResNet18 for classification task of Histopathologic Cancer Detection dataset using four different distillation techniques: ML - Mutual Learning, KD (on) - online Knowledge Distillation, KD (off) - offline Knowledge Distillation, and KD + ML – combined KD \& ML; and three different learning strategies -  V1 (predictions only), V2 (features only) and a diverse knowledge paradigm, V3 (both predictions and features).}
   
% \vspace{2.5mm}
\begin{subtable}{\linewidth}
    \centering
\resizebox{12.2cm}{!}{ 
            \begin{tabular}{|c|c|c|c|}
        \hline
 \multirow{5}*{KD(off)}&{V1} & V2 & V3   \\
         
         & ($\alpha=0.2,\alpha'=0.2$)   & ($\alpha=0.2,\alpha'=0.2$) & (${\alpha=0.2},\alpha'=0.2$)\\
         & ($\beta=0.8,\beta'=0.8$)   & ($\beta=0.8,\beta'=0.8$)& (${\beta=0.8},\beta'=0.8$)  \\
         & ($\gamma=0,\gamma'=0$)   & ($\gamma=0,\gamma'=0$) & (${\gamma=0},\gamma'=0$)\\
         \hline

\textbf{S1} &$95.82\pm0.106$  &$95.19\pm0.183$ &$95.3\pm0.141$ \\
\hline
\textbf{S2} &$95.57\pm0.106$ &$95.67\pm0.176$ &$95.47\pm0.247$ \\
\hline
\textbf{Ensemble} &$95.82\pm0.212$  &$95.55\pm0.637$ &${95.95\pm0.141}$ \\
 \hline
 \hline
         
    \end{tabular}
   } 
       \label{tab:52}
    \end{subtable}{}
    \begin{subtable}{\linewidth}
    \centering
\resizebox{12.2cm}{!}{ 
            \begin{tabular}{|c|c|c|c|}
        \hline
         \multirow{5}*{KD(on)}&{V1} & V2 & V3   \\
          & ($\alpha=0.2,\alpha'=0.2$)   & ($\alpha=0.2,\alpha'=0.2$) & (${\alpha=0.1},\alpha'=0.2$)\\
         & ($\beta=0.8,\beta'=0.8$)   & ($\beta=0.8,\beta'=0.8$)& (${\beta=0.9},\beta'=0.8$)  \\
         & ($\gamma=0,\gamma'=0$)   & ($\gamma=0,\gamma'=0$) & (${\gamma=0},\gamma'=0$)\\
         \hline
      
\textbf{S1} &$95.42\pm0.367$ &$95.44\pm0.961$ &$96.03\pm0.466$  \\
 \hline
\textbf{S2} &$95.14\pm0.121$ &$95.89\pm0.466$  &$95.47\pm0.339$\\
 \hline
\textbf{Ensemble} &$ 95.92\pm0.311$ &$96.01\pm0.480$ &${96.52\pm0.207} $ \\
\hline
 \hline
         
    \end{tabular}
    }
    \label{tab:53}
    \end{subtable}{}

    % \vspace{2.5mm}

    % \vspace{2.5mm}
    \begin{subtable}{\linewidth}
    \centering
\resizebox{12.2cm}{!}{ 
            \begin{tabular}{|c|c|c|c|}
        \hline
         \multirow{5}*{KD + ML}&{V1} & V2 & V3   \\
         & ($\alpha=0.1,\alpha'=0.2$)   & ($\alpha=0.2,\alpha'=0.2$) & ($\alpha=0.2,\alpha'=0.4$)\\
         & ($\beta=0.45,\beta'=0.4$)   & ($\beta=0.4,\beta'=0.4$)& ($\beta=0.4,\beta'=0.3$)  \\
         & ($\gamma=0.45,\gamma'=0.4$)   & ($\beta=0.4,\beta'=0.4$) & ($\gamma=0.4,\gamma'=0.3$)\\
         
         \hline

\textbf{S1} &$94.05\pm0.848$ &$95.71\pm0.466$ &$96.23\pm0.636$ \\
\hline
\textbf{S2} &$93.92\pm1.308$ &$95.98\pm0.028$ &$95.58\pm0.608$  \\
\hline
\textbf{Ensemble} &$94.60\pm1.272 $&$96.5\pm0.231 $ &${{96.41\pm0.777}}$ \\
\hline

 \hline
         
    \end{tabular}
    }
       \label{tab:54}
    \end{subtable}{}

\label{tab:5}
\end{table}

\begin{table*}[htbp]
% \captionsetup{labelfont=bf}
\renewcommand{\arraystretch}{1.2}
% \captionsetup{font=small}
\centering
        \caption{Fidelity comparison of U-Net with ResNet50-ResNet18 encoder for segmentation task using LGG dataset with IoU and F-score metrics using four different distillation techniques: ML - Mutual Learning, KD (on) - online Knowledge Distillation, KD (off) - offline Knowledge Distillation, and KD + ML – combined KD \& ML; and three different learning strategies -  V1 (predictions only), V2 (features only) and a diverse knowledge paradigm, V3 (both predictions and features).}
   
     \begin{subtable}{\linewidth}
     \centering
\resizebox{15cm}{!}{       
        \begin{tabular}{|c|c|c|c|c|c|c|}
        \hline
{KD} &\multicolumn{2}{c|} {V1} &\multicolumn{2}{c|} {V2} &\multicolumn{2}{c|} {V3} \\
        (off) & \multicolumn{2}{c|}{$\alpha=0.2,\alpha'=0.2$}  & \multicolumn{2}{c|}{$\alpha=0.2,\alpha'=0.2$} & \multicolumn{2}{c|}{$\alpha=0.2,\alpha'=0.2$} \\ 
         & \multicolumn{2}{c|}{$\beta=0.8,\beta'=0.8$}  
         & \multicolumn{2}{c|}{$\beta=0.8,\beta'=0.8$} 
         & \multicolumn{2}{c|}{$\beta=0.8,\beta'=0.8$} \\ 
         & \multicolumn{2}{c|}{$\gamma=0,\gamma'=0$}  & \multicolumn{2}{c|}{$\gamma=0,\gamma'=0$} & \multicolumn{2}{c|}{$\gamma=0,\gamma'=0$} \\ 
         \hline
         
&IoU  &F-score &IoU &F-score &IoU &F-score \\
\hline

\textbf{S1} &$81.97\pm1.681$ &$80.57\pm1.697$  &$83.94\pm0.968$ &$91.27\pm0.565$ &$82.17\pm0.205$ &$90.21+0.110$\\
\hline
\textbf{S2} &$82.19\pm0.779$ &$89.99\pm1.414$ &$82.85\pm1.315$ &$90.51\pm0.933$ &$82.28+0.813$ &$90.28+0.480$ \\
\hline
\textbf{Ensemble} &$83.74\pm1.800$ & $90.34\pm1.244$&$85.25\pm0.958$ &$92.05\pm0.551$  &$83.98\pm0.277$ &$90.67\pm0.629$  \\
\hline
 \hline
         
    \end{tabular}
    }
    
    \label{tab:62}
    \end{subtable}
% \vspace{2.5mm}
% **********************************************************************************
     \begin{subtable}{\linewidth}
     \centering
\resizebox{15cm}{!}{       
        \begin{tabular}{|c|c|c|c|c|c|c|}
        \hline
{KD} &\multicolumn{2}{c|} {V1} &\multicolumn{2}{c|} {V2} &\multicolumn{2}{c|} {V3} \\
         (on)& \multicolumn{2}{c|}{$\alpha=0.1,\alpha'=0.1$}  & \multicolumn{2}{c|}{$\alpha=0.2,\alpha'=0.2$} & \multicolumn{2}{c|}{$\alpha=0.1,\alpha'=0.2$} \\ 
         & \multicolumn{2}{c|}{$\beta=0.9,\beta'=0.9$}  
         & \multicolumn{2}{c|}{$\beta=0.8,\beta'=0.8$} 
         & \multicolumn{2}{c|}{$\beta=0.9,\beta'=0.8$} \\ 
         & \multicolumn{2}{c|}{$\gamma=0,\gamma'=0$}  & \multicolumn{2}{c|}{$\gamma=0,\gamma'=0$} & \multicolumn{2}{c|}{$\gamma=0,\gamma'=0$} \\ 
         \hline
         
&IoU  &F-score &IoU &F-score &IoU &F-score \\
 \hline
\textbf{S1} &$80.89\pm0.431$ &$89.44\pm0.268$  &$80.89\pm0.014$ &$89.44\pm0.021$ &$80.53\pm1.043$ &$89.20\pm1.251$\\ \hline
\textbf{S2} &$79.43\pm0.070$ &$88.53\pm0.049$ &$81.62\pm0.084$ &$89.97\pm0.049$ &$80.85\pm0.799$ &$89.43\pm0.480$ \\ \hline
\textbf{Ensemble} &$79.79\pm1.732$ &$88.75\pm1.080$ &$81.04\pm0.608$ &$89.52\pm0.367$ &$79.80\pm0.388$ &$89.26\pm0.473$ \\
\hline
 \hline
         
    \end{tabular}
    }
    
    \label{tab:63}
    \end{subtable}
% \vspace{1mm}
% **********************************************************************************

     \begin{subtable}{\linewidth}
     \centering
\resizebox{15cm}{!}{       
        \begin{tabular}{|c|c|c|c|c|c|c|}
        \hline
\multirow{5}*{KD + ML} &\multicolumn{2}{c|} {V1} &\multicolumn{2}{c|} {V2} &\multicolumn{2}{c|} {V3} \\
         & \multicolumn{2}{c|}{$\alpha=0.2,\alpha'=0.2$}  & \multicolumn{2}{c|}{$\alpha=0.2,\alpha'=0.2$} & \multicolumn{2}{c|}{$\alpha=0.1,\alpha'=0.1$} \\ 
         & \multicolumn{2}{c|}{$\beta=0.4,\beta'=0.4$}  & \multicolumn{2}{c|}{$\beta=0.4,\beta'=0.4$} & \multicolumn{2}{c|}{$\beta=0.45,\beta'=0.45$} \\ 
         & \multicolumn{2}{c|}{$\gamma=0.4,\gamma'=0.4$}  & \multicolumn{2}{c|}{$\gamma=0.4,\gamma'=0.4$} & \multicolumn{2}{c|}{$\gamma=0.45,\gamma'=0.45$} \\ 
         \hline
         
&IoU  &F-score &IoU &F-score &IoU &F-score \\

\hline
\textbf{S1} &$80.22\pm2.660$ &$89.01\pm1.640$ &$82.87\pm1.280$ &$90.62\pm0.770$ &$82.14\pm0.134$ &$90.19\pm0.077$  \\ \hline
\textbf{S2} &$80.60\pm1.746$ &$89.25\pm1.074$  &$81.82\pm0.183 $ &$90.06\pm0.113$ &$82.62\pm0.905$ &$90.48\pm0.537$ \\ \hline
\textbf{Ensemble} &$79.97\pm2.19$ &$88.23\pm1.994$  &$82.57\pm1.060$ &$90.45\pm0.636$ &$81.60\pm0.077$ &$89.93\pm0.134$\\
\hline
 \hline
         
    \end{tabular}
    }
    
    \label{tab:64}
    \end{subtable}
% \vspace{2.5mm}

\label{tab:6}
\end{table*}

% %%%%%%%%%%%%%%%%%%%%%%%%%%%%%%%%%%%%%%%%%%%%%%%%%%%%%%%%%%%%%%%%%%%%%%%%%%%%%%%%%%%%%%

\begin{figure*}[htbp]
\centering
% \captionsetup{labelfont=bf}
\includegraphics[width=0.8\linewidth]{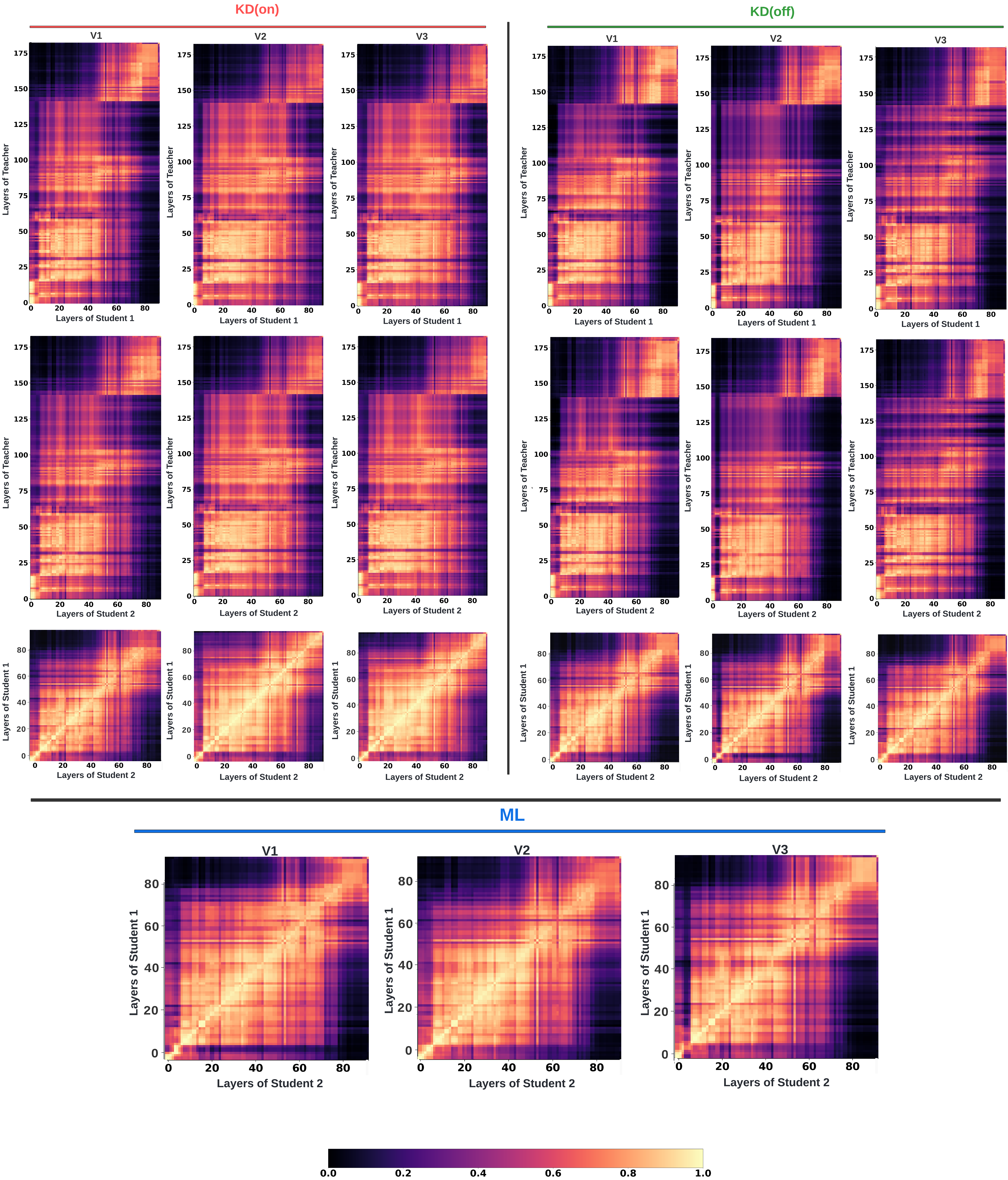}
\caption{CKA plots visualizing the similarity of learned representation between different layers of teacher and student networks using KD (on), KD (off), and ML models.}
\label{fig:8}
\end{figure*}

\begin{figure*}[htbp]
\centering
% \captionsetup{labelfont=bf}
\includegraphics[width=\linewidth]{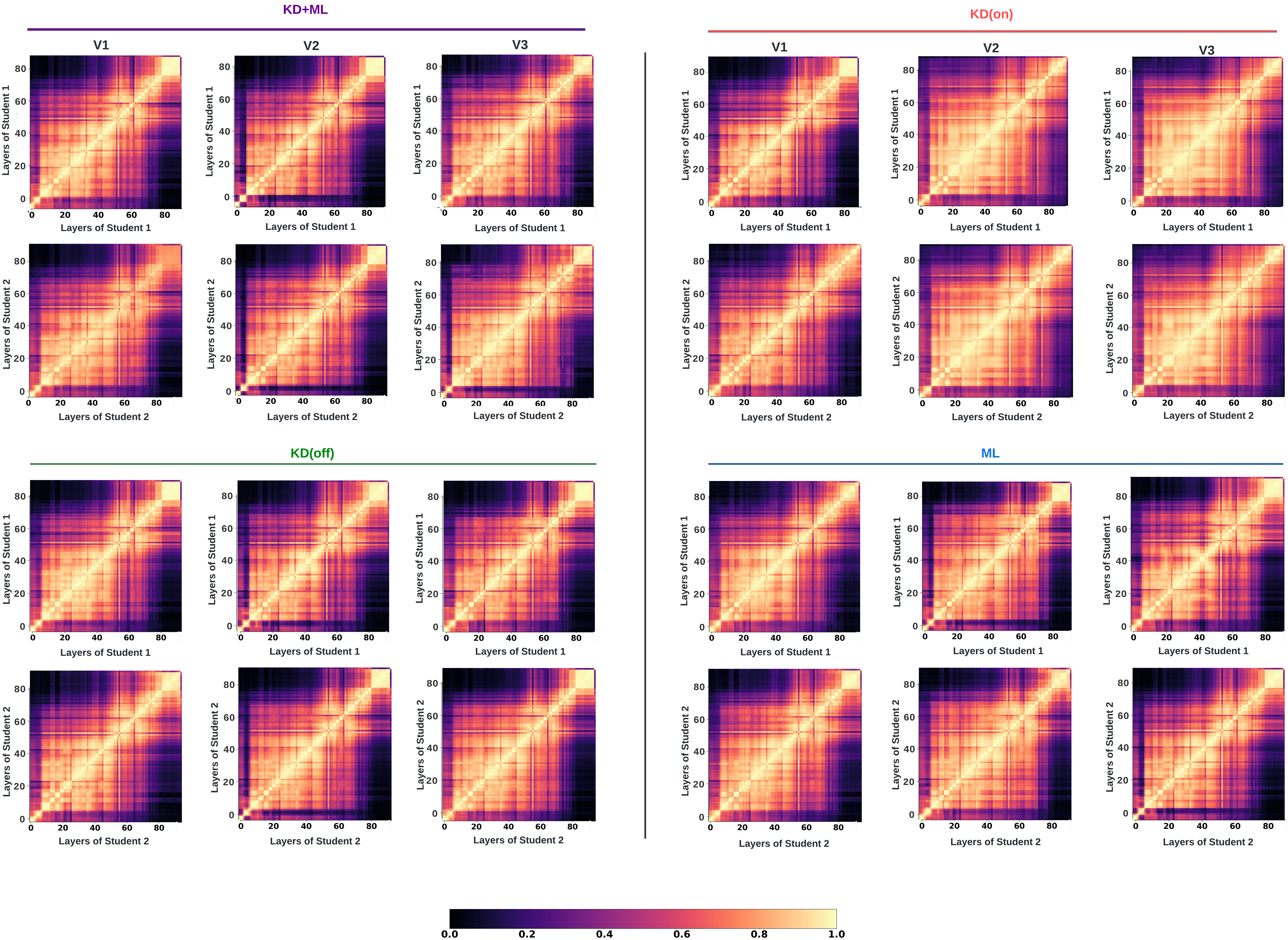}
\caption{CKA plots visualizing the similarity of learned representation within different layers of student networks using KD+ML, KD (on), KD (off), and ML models.}
\label{fig:9}
\end{figure*}

\clearpage
%%%%%%%%%%%%%%%%%%%%%%%%%%%%%%%%%%%%%%%%%%%%%%%%%%%%%%%%%%%%%%%%%%%%%%%%%%%%%%%%%%%%%%
% \colorlet{shadecolor}{yellow}
% \begin{shaded*}
\subsection{Classification dataset: HAM10000}
To rigorously evaluate the effectiveness of our proposed approach, we conducted a comprehensive series of experiments using the HAM10000 dataset \cite{ham}. This dataset comprises 10,015 images distributed across seven distinct classes and was originally sourced from \cite{oham}. Each image has a resolution
of $450\times600$ downsampled to $224\times224$. To ensure a fair and robust evaluation, we partitioned the dataset into training, validation, and test sets, closely mirroring the original distribution to maintain the dataset's integrity. We addressed the class imbalance challenge by employing a balanced combination of under-sampling and over-sampling techniques exclusively on the training data. 
\subsubsection{Evaluation metrics}
To ensure a fair evaluation of the multiclass HAM10000 classification, in addition to accuracy, we also used precision and recall, defined below:

\begin{equation}
Precision= \frac{TP}{TP+FP}
\end{equation}

\begin{equation}
Recall = \frac{TP}{TP+FN}
\end{equation}

Where TP, TN, FP, and FN represent the number of True Positive, True Negative, False Positive, and False Negative predictions, respectively.
\subsection{Segmentation dataset: HAM10000}
In the context of the segmentation task, we conducted an evaluation of our approach using the HAM10000 dataset, as referenced in \cite{ham}. The segmentation masks for this dataset were sourced from \cite{seg_dataset2} and \cite{ham_seg}. It's worth noting that we maintained consistent preprocessing procedures with those employed in the classification task.

% \end{shaded*}
%############################## 
% Baseline table
%#####################################
\begin{table}[h]
% \captionsetup{labelfont=bf}
\centering
% \captionsetup{font=small}
\caption{Performance comparison of baseline models for HAM10000 classification and segmentation tasks.}

% \resizebox{8.2cm}{!}{
\begin{tabular}{|c|c|c|c||c|c|c|}

\hline
\multirow{2}*{Model}&\multicolumn{3}{c|}{Classification} &\multicolumn{2}{c|}{{Segmentation}} \\
\cmidrule{2-6}
 & {Precision} & {Recall} &{Accuracy} &{IOU} &{F-score} \\
\hline
ResNet50 &$70.64\pm0.993$ & $72.48\pm0.404$ & $83.63\pm0.065$ &$87.77\pm0.339$ &$93.34\pm0.261$\\
\hline
ResNet18 & $65.57\pm0.249$ & $70.95\pm0.335$ & $82.08\pm0.291$ & $ 85.30\pm0.228 $& $ 89.88\pm0.118 $\\
\hline

\end{tabular}
% }
\label{tab:22}
\end{table}
% \begin{sidewaystable}
\begin{table*}[htbp]
\captionsetup{labelfont=bf}
\renewcommand{\arraystretch}{1.2}
\captionsetup{font=small}
\centering
        \caption{Performance comparison for Classification task using HAM10000 with Precision, Recall and Accuracy metrics using four different network configurations: ML - Mutual Learning, KD (on) - online Knowledge Distillation, KD (off) - offline Knowledge Distillation, and KD + ML – combined KD \& ML; and three different learning strategies - V1 (predictions only), V2 (features only) and a diverse knowledge paradigm, V3 (both predictions and features).}
    \begin{subtable}{\linewidth}
    \centering
    	% \hspace*{-100cm} 
\resizebox{16cm}{!}{       
        \begin{tabular}{|c|c|c|c|c|c|c|c|c|c|}
        \hline

\multirow{5}*{ML} &\multicolumn{3}{c|} {V1} &\multicolumn{3}{c|} {V2} &\multicolumn{3}{c|} {V3} \\
         & \multicolumn{3}{c|}{$\alpha=0.1,\alpha'=0.1$}  & \multicolumn{3}{c|}{$\alpha=0.2,\alpha'=0.2$} & \multicolumn{3}{c|}{$\alpha=0.2,\alpha'=0.4$} \\ 
         & \multicolumn{3}{c|}{$\beta=0,\beta'=0$}  
         & \multicolumn{3}{c|}{$\beta=0,\beta'=0$} 
         & \multicolumn{3}{c|}{$\beta=0,\beta'=0$} \\ 
         & \multicolumn{3}{c|}{$\gamma=0.9,\gamma'=0.9$}  & \multicolumn{3}{c|}{$\gamma=0.8,\gamma'=0.8$} & \multicolumn{3}{c|}{$\gamma=0.8,\gamma'=0.6$} \\ 
\hline
         
&Precision  &Recall  & Accuracy &Precision  &Recall  & Accuracy&Precision  &Recall  & Accuracy \\
\hline
\textbf{S1} &$66.03\pm0.636$ &$68.23\pm0.238$  &$81.78\pm0.912$ &$65.18\pm0.937$ &$72.63\pm0.449$ &$82.19\pm0.854$ & $67.46\pm1.356$ & $72.06\pm1.334$ & $83.21\pm0.415$\\
\hline
\textbf{S2} &$65.14\pm0.238$ &$71.27\pm1.097$  &$82.78\pm0.778$ &$66.80\pm0.945$ &$70.24\pm0.516$ &$81.87\pm0.864$ & $65.41\pm0.811$ & $71.74\pm0.276$ & $82.30\pm0.739$\\
\hline
\textbf{Ensemble} &$67.78\pm1.119$ &$72.46\pm1.986$ &$83.15\pm1.025$ &$67.44\pm1.098$ &$73.14\pm 0.988$&$82.33\pm0.138$  &$\mathbf{69.28\pm0.190}$ &$\mathbf{74.19\pm0.395}$ &$\mathbf{83.57\pm 0.095}$\\
\hline
 \hline
         
    \end{tabular}
    }
    
    \label{tab:91}
    \end{subtable}
% \vspace{2.5mm}
% **********************************************************************************
     \begin{subtable}{\linewidth}
     \centering
\resizebox{16cm}{!}{       
         \begin{tabular}{|c|c|c|c|c|c|c|c|c|c|}
        \hline

{KD} &\multicolumn{3}{c|} {V1} &\multicolumn{3}{c|} {V2} &\multicolumn{3}{c|} {V3} \\
        (off)   & \multicolumn{3}{c|}{$\alpha=0.2,\alpha'=0.2$}  & \multicolumn{3}{c|}{$\alpha=0.2,\alpha'=0.2$} & \multicolumn{3}{c|}{$\alpha=0.1,\alpha'=0.2$} \\ 
         & \multicolumn{3}{c|}{$\beta=0.8,\beta'=0.8$}  
         & \multicolumn{3}{c|}{$\beta=0.8,\beta'=0.8$} 
         & \multicolumn{3}{c|}{$\beta=0.9,\beta'=0.8$} \\ 
         & \multicolumn{3}{c|}{$\gamma=0,\gamma'=0$}  & \multicolumn{3}{c|}{$\gamma=0,\gamma'=0$} & \multicolumn{3}{c|}{$\gamma=0,\gamma'=0$} \\ 
    
         \hline
         
&Precision  &Recall  & Accuracy &Precision  &Recall  & Accuracy&Precision  &Recall  & Accuracy \\
\hline
\textbf{T} &$70.64\pm0.993$ & $72.48\pm0.404$ & $83.63\pm0.065$ &$70.64\pm0.993$ & $72.48\pm0.404$ & $83.63\pm0.065$ & $70.64\pm0.993$ & $72.48\pm0.404$ & $83.63\pm0.065$\\
\hline
\textbf{S1} &$67.10\pm1.513$ &$71.92\pm0.950$  &$81.81\pm0.732$ &$67.78\pm0.844$ &$70.79\pm0.844$ &$81.99\pm0.522$ & $69.96\pm0.735$ & $70.49\pm0.587$ & $82.88\pm0.180$\\
\hline
\textbf{S2} &$66.62\pm1.419$ &$71.17\pm0.274$  &$81.30\pm0.343$ &$68.42\pm0.744$ &$69.42\pm0.744$ &$81.64\pm0.663$ & $67.04\pm0.637$ & $73.04\pm0.556$ & $82.62\pm0.535$\\
\hline
\textbf{Ensemble} &$68.24\pm0.960$ &$72.07\pm0.461$ &$82.62\pm1.003$ &$69.08\pm0.499$ &$71.58\pm 0.499$&$82.25\pm0.443$  &$\mathbf{71.35\pm0.289}$ &$\mathbf{73.35\pm0.550}$ &$\mathbf{83.33\pm 0.190}$\\
\hline
 \hline
         
    \end{tabular}
    }
    
    \label{tab:92}
    \end{subtable}
% \vspace{2.5mm}
% **********************************************************************************
     \begin{subtable}{\linewidth}
     \centering
\resizebox{16cm}{!}{       
                 \begin{tabular}{|c|c|c|c|c|c|c|c|c|c|}
        \hline

{KD} &\multicolumn{3}{c|} {V1} &\multicolumn{3}{c|} {V2} &\multicolumn{3}{c|} {V3} \\
        (on)   & \multicolumn{3}{c|}{$\alpha=0.2,\alpha'=0.2$}  & \multicolumn{3}{c|}{$\alpha=0.2,\alpha'=0.2$} & \multicolumn{3}{c|}{$\alpha=0.1,\alpha'=0.2$} \\ 
         & \multicolumn{3}{c|}{$\beta=0.8,\beta'=0.8$}  
         & \multicolumn{3}{c|}{$\beta=0.8,\beta'=0.8$} 
         & \multicolumn{3}{c|}{$\beta=0.9,\beta'=0.8$} \\ 
         & \multicolumn{3}{c|}{$\gamma=0,\gamma'=0$}  & \multicolumn{3}{c|}{$\gamma=0,\gamma'=0$} & \multicolumn{3}{c|}{$\gamma=0,\gamma'=0$} \\     
         \hline
         
&Precision  &Recall  & Accuracy &Precision  &Recall  & Accuracy&Precision  &Recall  & Accuracy \\
\hline
\textbf{T} &$69.45\pm0.390$ &$70.22\pm0.577$  &$83.24\pm1.180$ &$71.48\pm0.997$ &$70.90\pm0.327$ &$82.06\pm0.725$ & $70.26\pm0.729$ & $70.48\pm0.514$ & $83.42\pm1.380$\\
\hline
\textbf{S1} &$69.16\pm0.513$ &$70.38\pm0.907$  &$81.72\pm0.997$ &$68.05\pm0.793$ &$72.81\pm0.543$ &$82.09\pm0.997$ & $69.10\pm0.848$ & $72.49\pm0.787$ & $82.24\pm0.374$\\
\hline
\textbf{S2} &$65.62\pm1.734$ &$72.14\pm0.416$  &$82.35\pm1.106$ &$67.42\pm0.424$ &$72.30\pm0.814$ &$82.87\pm0.519$ & $69.87\pm1.195$ & $71.51\pm0.652$ & $83.51\pm0.816$\\
\hline
\textbf{Ensemble} &$68.86\pm1.600$ &$73.61\pm0.466$ &$83.78\pm1.154$ &$69.24\pm0.512$ &$73.23\pm 0.586$&$83.16\pm0.636$  &$\mathbf{72.51\pm1.166}$ &$\mathbf{74.86\pm0.555}$ &$\mathbf{84.42\pm 0.364}$\\
\hline
 \hline
         
    \end{tabular}
    }
    
    \label{tab:93}
    \end{subtable}
% \vspace{1mm}
% **********************************************************************************

     \begin{subtable}{\linewidth}
     \centering
\resizebox{16cm}{!}{       
         \begin{tabular}{|c|c|c|c|c|c|c|c|c|c|}
        \hline
\multirow{5}*{KD + ML} &\multicolumn{3}{c|} {V1} &\multicolumn{3}{c|} {V2} &\multicolumn{3}{c|} {V3} \\
         & \multicolumn{3}{c|}{$\alpha=0.1,\alpha'=0.2$}  & \multicolumn{3}{c|}{$\alpha=0.2,\alpha'=0.2$} & \multicolumn{3}{c|}{$\alpha=0.2,\alpha'=0.4$} \\ 
         & \multicolumn{3}{c|}{$\beta=0.45,\beta'=0.4$}  & \multicolumn{3}{c|}{$\beta=0.4,\beta'=0.4$} & \multicolumn{3}{c|}{$\beta=0.4,\beta'=0.3$} \\ 
         & \multicolumn{3}{c|}{$\gamma=0.45,\gamma'=0.4$}  & \multicolumn{3}{c|}{$\gamma=0.4,\gamma'=0.4$} & \multicolumn{3}{c|}{$\gamma=0.4,\gamma'=0.3$} \\ 
         \hline
         
&Precision  &Recall  & Accuracy &Precision  &Recall  & Accuracy&Precision  &Recall  & Accuracy \\
\hline
\textbf{T} &$70.33\pm1.162$ &$72.96\pm1.186$  &$83.69\pm0.274$ &$68.12\pm0.814$ &$72.16\pm0.289$ &$82.60\pm1.154$ & $69.49\pm0.782$ & $72.71\pm0.569$ & $84.78\pm0.903$\\
\hline
\textbf{S1} &$68.33\pm1.256$ &$71.12\pm0.886$  &$83.33\pm0.261$ &$67.59\pm0.509$ &$74.57\pm0.879$ &$83.25\pm0.476$ & $72.15\pm0.601$ & $75.11\pm0.306$ & $83.69\pm0.113$\\
\hline
\textbf{S2} &$68.65\pm0.923$ &$73.74\pm0.793$  &$83.29\pm0.590$ &$67.88\pm0.262$ &$71.87\pm0.347$ &$82.43\pm1.085$ & $73.71\pm0.796$ & $74.88\pm1.516$ & $84.03\pm0.226$\\
\hline
\textbf{Ensemble} &$71.02\pm0.820$ &$75.75\pm0.583$  &$84.51\pm0.344$ &$69.35\pm0.505$ &$74.91\pm0.584$ &$83.87\pm0.481$ &\textcolor{blue}{\bm{$74.42\pm0.606$}} &\textcolor{blue}{\bm{$76.97\pm0.564$}}&\textcolor{blue}{\bm{$85.22\pm0.742$}}\\
\hline
 \hline
         
    \end{tabular}
    }
    
    \label{tab:94}
    \end{subtable}
% \vspace{2.5mm}

\label{tab:99}
\end{table*}
% \end{sidewaystable}
%############################## 
% Segmentation table
%##################################
\begin{table*}[t]
% \captionsetup{labelfont=bf}
\renewcommand{\arraystretch}{1.2}
% \captionsetup{font=small}
\centering
        \caption{Performance comparison of U-Net with ResNet50-ResNet18 encoder for segmentation task using HAM10000 dataset with IoU and F-score metrics using four different distillation techniques: ML - Mutual Learning, KD (on) - online Knowledge Distillation, KD (off) - offline Knowledge Distillation, and KD + ML – combined KD \& ML; and three different learning strategies -  V1 (predictions only), V2 (features only) and a diverse knowledge paradigm, V3 (both predictions and features). }
    \begin{subtable}{\linewidth}
    \centering
    	% \hspace*{-100cm} 
\resizebox{13cm}{!}{       
        \begin{tabular}{|c|c|c|c|c|c|c|}
        \hline
\multirow{5}*{ML} &\multicolumn{2}{c|} {V1} &\multicolumn{2}{c|} {V2} &\multicolumn{2}{c|} {V3} \\
         & \multicolumn{2}{c|}{$\alpha=0.2,\alpha'=0.2$}  & \multicolumn{2}{c|}{$\alpha=0.2,\alpha'=0.2$} & \multicolumn{2}{c|}{$\alpha=0.2,\alpha'=0.2$} \\ 
         & \multicolumn{2}{c|}{$\beta=0,\beta'=0$}  
         & \multicolumn{2}{c|}{$\beta=0,\beta'=0$} 
         & \multicolumn{2}{c|}{$\beta=0,\beta'=0$} \\ 
         & \multicolumn{2}{c|}{$\gamma=0.8,\gamma'=0.8$}  & \multicolumn{2}{c|}{$\gamma=0.8,\gamma'=0.8$} & \multicolumn{2}{c|}{$\gamma=0.8,\gamma'=0.8$} \\ 
         \hline
         
&IoU  &F-score &IoU &F-score &IoU &F-score \\
\hline
\textbf{S1} &$87.16\pm0.700$ &$93.10\pm0.431$ &$86.12\pm0.322$ &$92.57\pm0.325$ &$87.05\pm0.289$ &$93.64\pm0.162$  \\
\hline
\textbf{S2} &$87.06\pm0.346$ &$93.36\pm0.190$ &$86.20\pm0.583$ &$92.69\pm0.516$ &$87.12\pm0.275$ &$93.48\pm0.155$  \\
\hline
\textbf{Ensemble} &$87.36\pm0.054$ &$93.51\pm0.125$ &$86.78\pm0.516$ &$92.96\pm0.516$ &$\mathbf{88.05\pm0.410}$ &$\mathbf{93.73\pm0.233}$ \\
\hline
 \hline
         
    \end{tabular}
    }
    
    \label{tab:141}
    \end{subtable}
% \vspace{2.5mm}
% **********************************************************************************
     \begin{subtable}{\linewidth}
     \centering
\resizebox{13cm}{!}{       
        \begin{tabular}{|c|c|c|c|c|c|c|}
        \hline
{KD} &\multicolumn{2}{c|} {V1} &\multicolumn{2}{c|} {V2} &\multicolumn{2}{c|} {V3} \\
        (off)   & \multicolumn{2}{c|}{$\alpha=0.2,\alpha'=0.2$}  & \multicolumn{2}{c|}{$\alpha=0.2,\alpha'=0.2$} & \multicolumn{2}{c|}{$\alpha=0.2,\alpha'=0.2$} \\ 
         & \multicolumn{2}{c|}{$\beta=0.8,\beta'=0.8$}  
         & \multicolumn{2}{c|}{$\beta=0.8,\beta'=0.8$} 
         & \multicolumn{2}{c|}{$\beta=0.8,\beta'=0.8$} \\ 
         & \multicolumn{2}{c|}{$\gamma=0,\gamma'=0$}  & \multicolumn{2}{c|}{$\gamma=0,\gamma'=0$} & \multicolumn{2}{c|}{$\gamma=0,\gamma'=0$} \\ 
         \hline
         
&IoU  &F-score &IoU &F-score &IoU &F-score \\
\hline
\textbf{T} &$87.77\pm0.339$ &$93.34\pm0.261$  &$87.77\pm0.339$ &$93.34\pm0.261$  &$87.77\pm0.339$ &$93.34\pm0.261$ \\
\hline
\textbf{S1} &$87.32\pm0.254$ &$93.24\pm0.156$ &$86.83\pm0.247$ &$93.01\pm0.565$ &$86.62\pm1.064$ &$92.82\pm1.187$ \\
\hline
\textbf{S2} &$87.11\pm0.480$ & $93.40\pm0.275$&$86.73\pm0.063$ &$93.06\pm0.035$  &$87.93\pm0.299$ &$93.28\pm0.452$  \\
\hline
\textbf{Ensemble} &$87.44\pm0.226$ &$93.66\pm0.127$&$87.06\pm0.363$ &$93.15\pm0.253$  &$\mathbf{87.65\pm0.728}$ &$\mathbf{93.58\pm0.028}$ \\
\hline
 \hline
         
    \end{tabular}
    }
    
    \label{tab:142}
    \end{subtable}
% \vspace{2.5mm}
% **********************************************************************************
     \begin{subtable}{\linewidth}
     \centering
\resizebox{13cm}{!}{       
        \begin{tabular}{|c|c|c|c|c|c|c|}
        \hline
{KD} &\multicolumn{2}{c|} {V1} &\multicolumn{2}{c|} {V2} &\multicolumn{2}{c|} {V3} \\
         (on)& \multicolumn{2}{c|}{$\alpha=0.1,\alpha'=0.1$}  & \multicolumn{2}{c|}{$\alpha=0.2,\alpha'=0.2$} & \multicolumn{2}{c|}{$\alpha=0.1,\alpha'=0.2$} \\ 
         & \multicolumn{2}{c|}{$\beta=0.9,\beta'=0.9$}  
         & \multicolumn{2}{c|}{$\beta=0.8,\beta'=0.8$} 
         & \multicolumn{2}{c|}{$\beta=0.9,\beta'=0.8$} \\ 
         & \multicolumn{2}{c|}{$\gamma=0,\gamma'=0$}  & \multicolumn{2}{c|}{$\gamma=0,\gamma'=0$} & \multicolumn{2}{c|}{$\gamma=0,\gamma'=0$} \\ 
         \hline
         
&IoU  &F-score &IoU &F-score &IoU &F-score \\
\hline
\textbf{T} &$87.55\pm0.247$ &$93.36\pm0.141$  &$86.76\pm0.636$ &$92.90\pm0.366$ &$87.32\pm0.417$ &$93.23\pm0.240$\\ \hline
\textbf{S1} &$87.66\pm0.558$ &$93.42\pm0.311$ &$87.32\pm0.494$ &$93.23\pm0.282$ &$86.86\pm1.499$ &$92.96\pm0.855$ \\ \hline
\textbf{S2} &$87.65\pm0.586$ &$93.42\pm0.332$ &$87.05\pm0.551$ &$93.07\pm0.311$ &$87.65\pm0.346$ &$93.42\pm0.197$ \\ \hline
\textbf{Ensemble} &$\mathbf{87.93\pm0.346}$ &$\mathbf{93.58\pm0.197}$ &$87.40\pm0.071$ &$93.28\pm0.042$ &${87.73\pm0.445}$ &${93.46\pm0.254}$ \\
\hline
 \hline
         
    \end{tabular}
    }
    
    \label{tab:143}
    \end{subtable}
% \vspace{1mm}
% **********************************************************************************

     \begin{subtable}{\linewidth}
     \centering
\resizebox{13cm}{!}{       
        \begin{tabular}{|c|c|c|c|c|c|c|}
        \hline
\multirow{5}*{KD + ML} &\multicolumn{2}{c|} {V1} &\multicolumn{2}{c|} {V2} &\multicolumn{2}{c|} {V3} \\
         & \multicolumn{2}{c|}{$\alpha=0.2,\alpha'=0.2$}  & \multicolumn{2}{c|}{$\alpha=0.2,\alpha'=0.2$} & \multicolumn{2}{c|}{$\alpha=0.1,\alpha'=0.1$} \\ 
         & \multicolumn{2}{c|}{$\beta=0.4,\beta'=0.4$}  & \multicolumn{2}{c|}{$\beta=0.4,\beta'=0.4$} & \multicolumn{2}{c|}{$\beta=0.45,\beta'=0.45$} \\ 
         & \multicolumn{2}{c|}{$\gamma=0.4,\gamma'=0.4$}  & \multicolumn{2}{c|}{$\gamma=0.4,\gamma'=0.4$} & \multicolumn{2}{c|}{$\gamma=0.45,\gamma'=0.45$} \\ 
         \hline
         
&IoU  &F-score &IoU &F-score &IoU &F-score \\
\hline
\textbf{T} &$87.39\pm0.975$ &$93.19\pm0.452$ &$87.45\pm0.671$ &$93.30\pm0.381$  &$88.12\pm0.979$ &$93.15\pm0.966$\\ \hline
\textbf{S1} &$88.08\pm0.183$ &$93.66\pm0.098$ &$87.64\pm1.096$ &$93.41\pm0.622$ &$88.86\pm1.004$ &$94.06\pm0.509$  \\ \hline
\textbf{S2} &$87.80\pm0.530$ &$93.50\pm0.304$  &$87.68\pm0.784 $ &$93.43\pm0.445$ &$88.91\pm0.954$ &$94.09\pm0.480$ \\ \hline
\textbf{Ensemble} &$88.37\pm0.311$ &$93.82\pm0.176$  &$87.98\pm0.776$ &$93.60\pm0.445$ &\textcolor{blue}{\bm{$89.40\pm0.799$}} &\textcolor{blue}{\bm{$94.27\pm0.700$}}\\
\hline
 \hline
         
    \end{tabular}
    }
    
    \label{tab:144}
    \end{subtable}
% \vspace{2.5mm}

\label{tab:115}
\end{table*}

\section*{Notes}
\printnotes*
\end{document}